\documentclass{article}

\PassOptionsToPackage{numbers,sort&compress,square}{natbib}

\usepackage[preprint]{neurips_2026}
\usepackage[table]{xcolor}
\usepackage[utf8]{inputenc}
\usepackage[T1]{fontenc}
\usepackage{graphicx}
\usepackage{booktabs}
\usepackage{multirow}
\usepackage{amsmath}
\usepackage{amsthm}
\usepackage[table]{xcolor}
\usepackage{adjustbox}
\usepackage{xspace}
\usepackage{microtype}
\usepackage{amssymb}
\usepackage{bm}
\usepackage{pifont}
\usepackage{hyperref}

\usepackage[table]{xcolor}

\definecolor{TickGreen}{RGB}{46,125,50}
\definecolor{CrossRed}{RGB}{198,40,40}
\newcommand{\cmark}{\textcolor{TickGreen}{\ding{51}}}
\newcommand{\xmark}{\textcolor{CrossRed}{\ding{55}}}
\usepackage{tikz}
\usetikzlibrary{positioning, calc}
\theoremstyle{plain}

\theoremstyle{remark}

\usepackage{color,soul}

\usepackage{hyperref}
\usepackage[capitalize,nameinlink]{cleveref}

\makeatletter
\DeclareRobustCommand\onedot{\futurelet\@let@token\@onedot}
\def\@onedot{\ifx\@let@token.\else.\null\fi\xspace}

\makeatother

\usepackage{pgfplots}
\usepackage{subcaption}
\usepackage{xcolor}
\pgfplotsset{compat=1.18}

\definecolor{unicon}{HTML}{B4C7E7}      %
\definecolor{human}{HTML}{2F5597}       %
\definecolor{finetune}{HTML}{A5A5A5}    %
\definecolor{parallel}{HTML}{7F7F7F}    %
\definecolor{unish}{HTML}{D9D9D9}       %

\newcommand{\modelname}{\mbox{UniCon3R}\xspace}

\newcommand{\supmat}{\textbf{Sup.~Mat.}\xspace}

\newcommand{\sota}{\mbox{state-of-the-art}\xspace}
\newcommand{\smplx}{\mbox{SMPL-X}\xspace}

\newcommand{\IVLM}{\mbox{InteractVLM}\xspace}
\newcommand{\DECO}{\mbox{DECO}\xspace}

\newcommand{\tabstyle}{%
  \footnotesize
  \setlength{\tabcolsep}{5pt}
  \renewcommand{\arraystretch}{1.18}
}

\definecolor{freeze_blue}{HTML}{1f78b4}
\definecolor{learn_red}{HTML}{d45c43}

\newcommand{\bu}{\mathbf{u}}

\begin{document}

\title{\modelname: Unified Contact-aware 4D Human-Scene Reconstruction from Monocular Video}

\author{%
\textbf{Tanuj Sur$^{1}$
\quad Shashank Tripathi$^{2}$
\quad Nikos Athanasiou$^{2}$} \\
\textbf{Ha Linh Nguyen$^{1}$
\quad Kai Xu$^{1}$
\quad Michael J.~Black$^{2}$
\quad Angela Yao$^{1}$} \\
$^{1}$National University of Singapore \\
$^{2}$Max Planck Institute for Intelligent Systems, T\"ubingen, Germany \\[0.3em]
\texttt{\{tanujsur,halinh,kxu,ayao\}@comp.nus.edu.sg} \\
\texttt{\{stripathi,nathanasiou,black\}@tuebingen.mpg.de}
}

\maketitle

\begin{abstract}

We introduce \textbf{\modelname}, a unified feed-forward framework for online human-scene 4D reconstruction from monocular video. Current feed-forward human-scene reconstruction methods suffer from artifacts, where bodies float above the ground or penetrate parts of the scene. A key reason is the lack of effective interaction modelling between the human and the environment. Our goal is to exploit contact between the human and the scene during inference to actively improve the human mesh reconstruction.
To that end, we explicitly model interaction by inferring 4D contact from the human pose and scene geometry and use the contact as a corrective cue for generating the pose. This enables \modelname to jointly recover scene  geometry and spatially aligned 4D humans within the scene. Experiments on standard human-centric video benchmarks show that \modelname\ outperforms \sota baselines on physical plausibility and global human motion estimation while preserving fast, feed-forward inference speeds. The results validate our central claim: contact serves as a powerful internal prior, thus establishing a new paradigm for physically grounded joint human-scene reconstruction. \textbf{ Source code and models will be released upon acceptance.}

\end{abstract}

\section{Introduction}
\label{sec:intro}

Recent work on 3D and 4D human-scene reconstruction~\cite{weng2021holistic, yalandur2025physic, muller2025reconstructing, chen2025human3r, JOSH3R, li2026unish, zhao2026onlinehmr} from monocular images and videos %
can recover human meshes and their surrounding environments. Unified feed-forward methods~\cite{chen2025human3r, li2026unish, zhao2026onlinehmr} are especially promising because they predict global human motion, dense scene geometry, and camera trajectories within a single forward pass, enabling online 4D reconstruction.
However, joint scene reconstruction and %
Human Mesh Recovery (HMR)~\cite{Multi-HMR, cai2023smpler, dwivedi2024tokenhmr, goel2023humans, tripathi2023ipman, kocabas2024pace, li2024coin, sun2022putting} in a %
world coordinate frame does not by itself guarantee physical plausibility~\cite{villegas2021contact, tripathi2024humos, yalandur2025physic}. Specifically, a body may be accurately localized in 3D yet still float above the floor, penetrate objects, or slide over supporting surfaces. In everyday human-scene interaction, physical coupling with the environment is mediated by contact: the body is supported by scene surfaces and interacts with objects through local contact regions.
We posit that the 3D contacts between the human and the scene provide important constraints that could improve the joint reconstruction.

 \begin{figure*}[t]
    \centering
    \includegraphics[width=\linewidth]{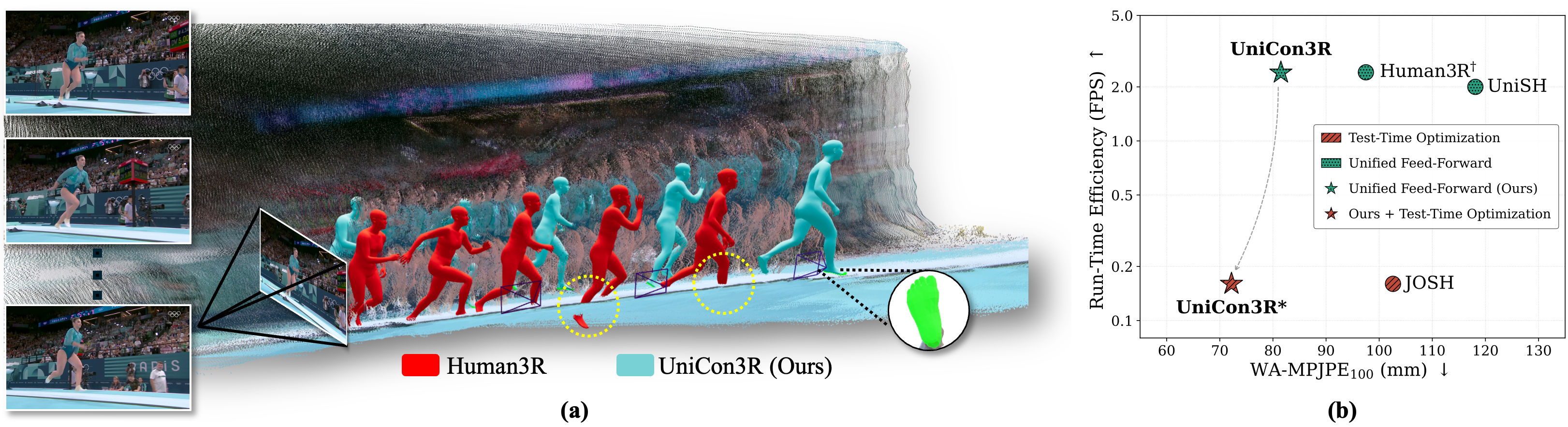}
    \caption{
    \textbf{Contact improves unified 4D human-scene reconstruction.}
    \textbf{(a)} Human3R~\cite{chen2025human3r} meshes are misaligned, %
    with the feet penetrating the ground (dashed yellow circles). \modelname\ predicts dense per-vertex contact (green inset) and uses it to improve body-scene alignment.
    \textbf{(b)} \textbf{Amortized runtime versus %
    global human motion  error %
    }. \modelname\ achieves the \sota on the RICH dataset~\cite{huang2022capturing}; %
    test-time optimization further reduces the error, but also reduces runtime efficiency.}

\label{fig:teaser}
\end{figure*}

Unfortunately, contact estimation is inherently ill-posed because the contact areas are often occluded in 2D images ~\cite{tripathi2023deco, dwivedi2025interactvlm, cseke2025pico}. %
Existing unified feed-forward  methods~\cite{chen2025human3r, li2026unish, JOSH3R} %
handle the physical interaction between the human and the scene %
implicitly, as the human reconstruction branch
does not leverage scene-grounded interaction cues during inference.
Resulting reconstructions exhibit artifacts such as floating feet, penetrations, or inconsistent dynamics, as shown in~\cref{fig:teaser}(a). This leaves a clear gap between merely reconstructing a person in a scene and %
physically plausible reconstructions.

\vspace{-2mm}

To address this gap, we introduce \textbf{\modelname}, a contact-aware framework for unified feed-forward~\footnote{Following CUT3R~\cite{cut3r}, we use \emph{feed-forward} to mean direct online inference without per-scene optimization or global alignment, even though the model maintains a recurrent persistent state.} 4D human-scene reconstruction. Our key insight is to treat contact as an internal feedback signal that interacts with both the human and scene reconstruction branches, enabling more accurate human localization within the forward pass. Building upon the prompting mechanisms of CUT3R~\cite{cut3r} and Human3R~\cite{chen2025human3r}, we introduce a \emph{scene-aware contact prompt} that gathers physical cues from the current frame, recurrent scene memory, and local scene geometry. The resulting contact prompt is decoded jointly with the human and scene tokens, allowing contact cues to influence the latent representation used for final human mesh reconstruction.

In principle, a model that predicts contact should help improve the body's world-frame placement~\cite{cseke2025pico,yalandur2025physic}.
However, we find that this alone is not sufficient. As our experiments show, such a model may correctly identify where contact is likely to occur, yet still fail to localize the body accurately. %
Thus, the key challenge is %
to ensure that predicted contact actively shapes the recovered human mesh within the unified feed-forward pipeline~\cite{yalandur2025physic, xue2024hsr}. 

To this end, rather than using contact as an auxiliary prediction~\cite{tripathi2023deco, dwivedi2025interactvlm}, we feed the inferred contact cues back into the HMR branch via a \emph{contact-guided latent refinement}. This allows the refined contact token to produce a residual update to the human latent representation during the forward pass. 
By internalizing contact, \modelname\ preserves the simplicity and streaming capabilities of unified feed-forward inference~\cref{fig:teaser}(b) while ensuring the recovered human mesh is physically plausible in scene-afforded constraints. Notably, incorporating contact into 4D human-scene reconstruction also boosts contact estimation accuracy, highlighting the strong synergy between the two tasks.

We evaluate \modelname\ on the challenging human-centric video benchmarks RICH~\cite{huang2022capturing}, EMDB~\cite{kaufmann2023emdb}, 3DPW~\cite{3DPW} and SLOPER4D~\cite{dai2023sloper4d}, focusing on physical plausibility, global human motion estimation and local human mesh recovery. Our results show that internalizing contact as a refinement signal reduces penetration by 
\textbf{74\%}, and improves WA-MPJPE metric by \textbf{16\%} on the RICH~\cite{huang2022capturing} dataset, while preserving online feed-forward inference.
In addition, on RICH~\cite{huang2022capturing}, UniCon3R improves contact recall over DECO~\cite{tripathi2023deco} while achieving higher F1 and lower geometric contact error.

In summary, our contributions are as follows:
    (i) We introduce a new design paradigm for unified feed-forward 4D human-scene reconstruction by modeling human-scene contact as an \emph{internal corrective signal}. %
    (ii) We propose \textbf{\modelname}, a novel contact-aware framework that internalizes human-scene interaction within unified feed-forward reconstruction through a \emph{scene-aware contact prompt} and \emph{contact-guided latent refinement}.
    (iii) We show that actively internalizing contact improves physical plausibility, global human motion and local human mesh recovery %
    over existing \sota feed-forward methods, while preserving competitive streaming efficiency.

\section{Related Work}
\label{sec:related}

\paragraph{3D Human Reconstruction} from monocular views 
has evolved from optimization-based fitting to feed-forward regression. Early methods such as SMPLify~\cite{bogo2016keep} fit parametric body models such as SMPL~\cite{SMPL} or SMPL-X~\cite{SMPLX} to image evidence. More recent HMR methods directly regress body parameters from images or videos~\cite{kanazawa2018end, li2022cliff, dwivedi2024tokenhmr, tripathi2023ipman, goel2023humans, xu2023smpler, cai2023smpler, CameraHMR, wang2023refit, yin2025smplest}, improving robustness and accuracy under challenging viewpoints, occlusions, and motion. Some recent methods also estimate global human motion in world coordinates from monocular video, but typically without explicitly modeling physical interaction with the surrounding 3D scene. In contrast, our goal is not only to recover a plausible human body, but to refine it within a jointly reconstructed metric scene, where scene-grounded contact acts as an internal corrective cue for the final human estimate.

\noindent\textbf{3D Human-Scene Contact}
provides direct evidence of how the body is supported by the environment. RICH~\cite{huang2022capturing} introduced dense full-body human-scene contact labels by combining multiview human capture with scanned 3D scenes, and proposed BSTRO for vertex-level body-scene contact estimation from images. DECO~\cite{tripathi2023deco} predicts dense 3D contact on the SMPL body from in-the-wild images using body-part and scene-context attention. InteractVLM~\cite{dwivedi2025interactvlm} estimates contact on both humans and objects by lifting 2D foundation-model predictions into 3D. PICO~\cite{cseke2025pico} further studies dense body-object contact correspondences and uses them for optimization-based human-object reconstruction. However, contact is typically treated either as a separate prediction target or as a signal used only in a later fitting stage. For example, JOSH~\cite{JOSH3R} uses BSTRO~\cite{huang2022capturing} as an off-the-shelf contact model to refine body pose during motion estimation. In contrast, we incorporate contact directly into the feed-forward reconstruction pathway.

\noindent\textbf{Joint Human-Scene Reconstruction}
recovers people together with the surrounding 3D environment. Early methods often rely on optimization and are restricted to static or multi-view settings~\cite{pavlakos2022one, muller2025reconstructing, rojas2025hamst3r}. Later monocular methods couple human priors with scene geometry. SyncHMR~\cite{zhao2024synergistic} uses SMPL-based constraints for metric-scale reconstruction, while JOSH~\cite{JOSH3R} initializes from dense scene reconstruction and human mesh recovery, then optimizes scene geometry, camera poses, and human motion with contact constraints. These methods can improve body-scene consistency, but require expensive %
per-scene optimizations.

Recent methods reduce this cost with feed-forward or optimization-free designs. JOSH3R~\cite{JOSH3R} distills JOSH pseudo-labels into an optimization-free model. %
Human3R~\cite{chen2025human3r} adapts CUT3R~\cite{cut3r} with visual prompt tuning to predict humans, scenes, and camera motion in a unified feed-forward model while keeping the scene backbone frozen. UniSH~\cite{li2026unish} combines scene-reconstruction and HMR priors to recover scene geometry, camera parameters, and metric-scale SMPL bodies in a single forward pass. These methods speed up joint reconstruction %
but human-scene interaction is still implicit, %
which often leads to physically implausible results. Our method models contact explicitly and feeds the contact representation back into the HMR pathway before the body regression.

\section{Preliminaries}
\label{sec:preliminaries}

\noindent\textbf{World-Grounded Motion \& 4D Scene Reconstruction.}
Modeling human-scene interaction requires representing the human and camera in a common world coordinate frame. In this work, we adopt the SMPL-X~\cite{SMPLX} human model and express the world-frame body vertices as
$\mathbf{V}^{\mathrm{body}}_t \in \mathbb{R}^{V \times 3}$,
with $\mathbf{V}^{\mathrm{body}}_t = \mathrm{SMPL\mbox{-}X}(\boldsymbol{\theta}_t,\boldsymbol{\beta},\boldsymbol{\alpha}_t,\mathbf{P}_t)$. Here, $\boldsymbol{\theta}_t$, $\boldsymbol{\beta}$, and $\boldsymbol{\alpha}_t$ denote the pose, shape, and expression parameters of the SMPL-X model. $\mathbf{P}^{\mathrm{cam}}_t$ is the local root translation in the camera frame and $\mathbf{T}_t$ is the camera pose that maps this local translation into the world frame.

\begin{figure*}[t]
    \centering
    \includegraphics[width=\linewidth]{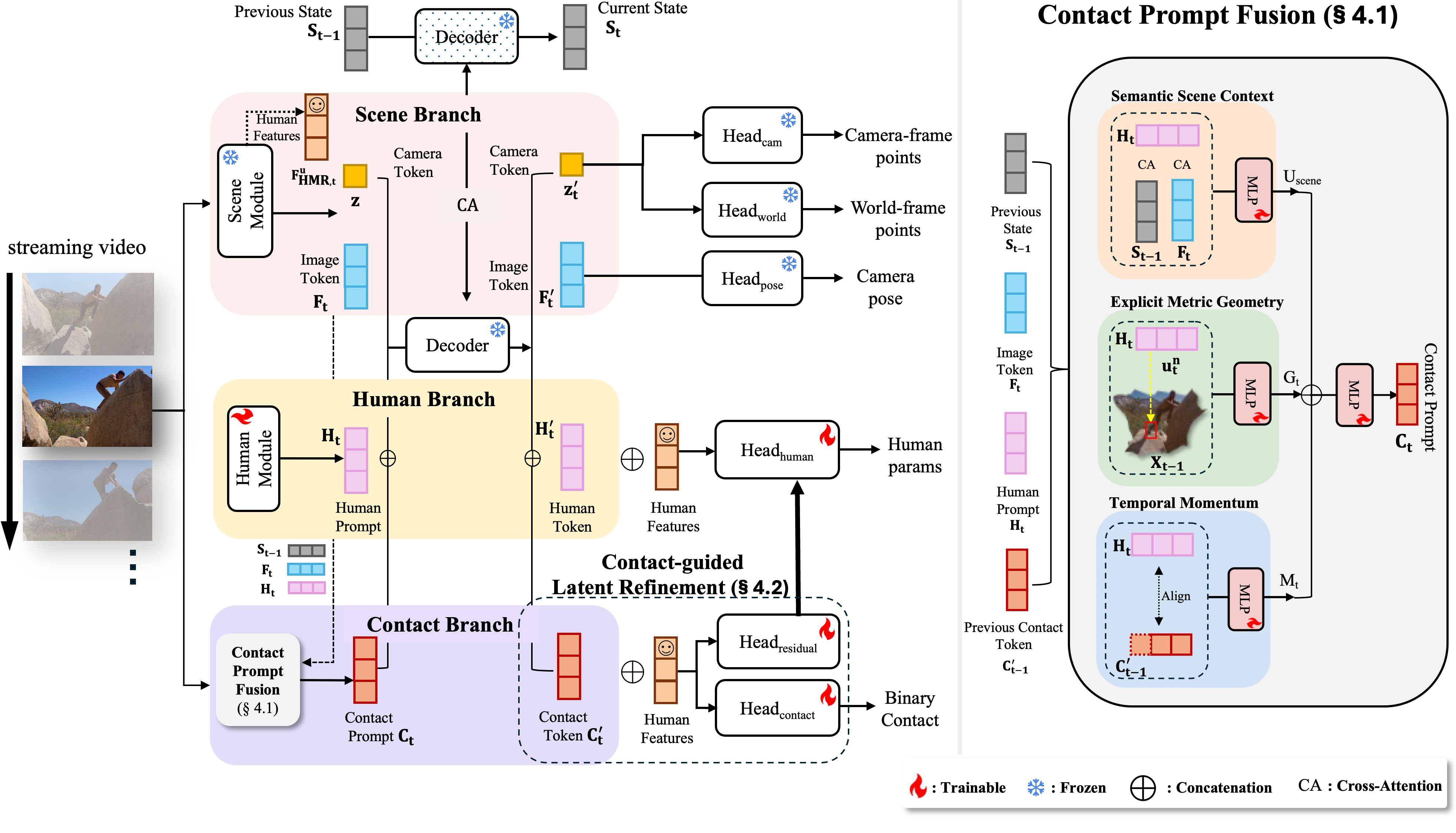}
    \caption{\textbf{Overview of \modelname.} \textbf{Left:} At time $t$, the Scene and Human branch encodes the current frame into image tokens $\mathbf{F}_t$ and a human prompt $\mathbf{H}_t$, to interact with the previous state $\mathbf{S}_{t-1}$ %
    and update the current state $\mathbf{S}_t$, to predict scene, camera, and human parameters. %
    Our contact branch outputs dense vertex-level contact on the SMPL~\cite{SMPL} mesh.  \textbf{Right:} \modelname\ constructs a \emph{scene-aware contact prompt} $\mathbf{C}_t$ (\S\ref{ssec:contact_prompt}) from semantic scene context, explicit metric geometry, and a temporal momentum term.  This gets decoded %
    into a refined contact token $\mathbf{C}_t'$ %
    which predicts dense SMPL contact.  It also drives \emph{contact-guided latent refinement} (\S\ref{ssec:latent_refinement}) by producing a residual update to the human latent before final SMPL-X regression.
    }
    
    \label{fig:pipeline}
\end{figure*}

Our scene reconstruction module builds on CUT3R's~\cite{cut3r} 4D scene reconstruction model. As shown in \cref{fig:pipeline} (top), given an image stream, it maintains a latent scene state $\mathbf{S}_t$ updated at each frame. Here, $t$ denotes the frame index. At time step $t$, the input image is encoded into feature tokens $\mathbf{F}_t$, which interact with the previous state $\mathbf{S}_{t-1}$ together with a learnable pose token $\mathbf{z}\in\mathbb{R}^{c}$, where $c$ is the decoder embedding dimension. This produces updated tokens $\mathbf{F}'_t$, $\mathbf{z}'_t$, and an updated state $\mathbf{S}_t$. Two separate heads decode camera-frame and world-frame scene pointmaps, $\mathbf{X}^{\mathrm{cam}}_t$ and $\mathbf{X}^{\mathrm{world}}_t$, and regresses the camera pose $\mathbf{T}_t$ from the updated tokens $\mathbf{F}'_t$, $\mathbf{z}'_t$.

\noindent\textbf{4D Human Reconstruction via visual prompting.}
\label{sec:human_branch_prelim}
For reconstructing the human mesh, we follow Human3R~\cite{chen2025human3r}, which extends CUT3R~\cite{cut3r} from pure scene reconstruction to joint human-scene reconstruction through parameter-efficient Visual Prompt Tuning (VPT)~\cite{vpt}. Rather than modifying the frozen CUT3R backbone, Human3R adds a human prompt $\mathbf{H}_t \in \mathbb{R}^{N_t \times c}$ where $N_t$ denotes the number of detected humans in frame $t$. {As shown in} \cref{fig:pipeline} (middle), we detect human anchors $\{\mathbf{u}_t^n\}_{n=1}^{N_t}$ from the CUT3R image tokens and gather the corresponding human-prior features $\mathbf{F}_{\mathrm{HMR},t}^{\bu}$ from a frozen Multi-HMR encoder~\cite{Multi-HMR}.

The human prompt in the HMR pathway acts as a human-specific query: it self-attends to image tokens to aggregate body evidence and cross-attends to the persistent scene state to retrieve temporally consistent human information anchored in the world-centered scene context.
The interaction can be summarized as:
$[\mathbf{F}'_t,\mathbf{z}'_t,\mathbf{H}'_t],\mathbf{S}_t = \textcolor{freeze_blue}{\mathrm{Decoders}}([\mathbf{F}_t,\mathbf{z},\mathbf{H}_t],\mathbf{S}_{t-1}),\; \mathbf{Y}_t = \textcolor{learn_red}{\mathrm{Head}_{\mathrm{human}}}(\mathbf{H}'_t)$
where $\mathbf{Y}_t$ denotes the predicted SMPL-X~\cite{SMPLX} parameters. Here, \textcolor{freeze_blue}{blue} denotes frozen modules inherited from CUT3R and Human3R, while \textcolor{learn_red}{red} denotes learnable modules introduced or fine-tuned in our framework. 
These preliminaries expose the key limitation of Human3R-style unified reconstruction: %
the interaction between $\mathbf{F}'_t$ and $\mathbf{H}_t$ cannot %
ensure physically grounded reconstruction.  %

\section{\modelname}
\label{sec:method}

\textbf{\modelname} is a contact-aware framework for online 4D human-scene reconstruction. %
Contact is encoded as an explicit prompt, processed by the same frozen decoder-and-memory pipeline used for scene and human reconstruction, and then used to refine the recovered body. The method has two components: \textbf{(1)} a scene-aware contact prompt that aggregates interaction evidence from the current frame, recurrent scene memory, and local scene geometry, and \textbf{(2)} a contact-guided latent refinement mechanism that feeds the refined contact representation back into the HMR pathway before final SMPL-X regression~\cite{SMPLX}. \cref{fig:pipeline} shows an overview of the framework.

\subsection{Contact Prompt Fusion}
\label{ssec:contact_prompt}

\noindent\textbf{Semantic Scene Context.}
 Robust contact reasoning requires immediate spatial evidence (e.g., observing a shoe hitting the floor). To recover such cues, we extract two complementary scene features,
$\mathbf{U}_{\mathrm{curr}} = \mathrm{CA}_{\mathrm{curr}}\!\left(\mathbf{H}_t, \mathbf{F}_t\right),\; \mathbf{U}_{\mathrm{mem}} = \mathrm{CA}_{\mathrm{mem}}\!\left(\mathbf{H}_t, \mathbf{S}_{t-1}\right)$ 
where $\mathrm{CA}(\cdot,\cdot)$ denotes cross-attention. 
Here, $\mathbf{U}_{\mathrm{curr}}$ captures visual evidence from the current frame, while $\mathbf{U}_{\mathrm{mem}}$ retrieves persistent object context from the recurrent scene state. We combine these cues through a learned gate:
$\boldsymbol{\gamma}_t = \sigma\!\left(\textcolor{learn_red}{\mathrm{MLP}}\!\left(\mathbf{H}_t \oplus \mathbf{U}_{\mathrm{curr}} \oplus \mathbf{U}_{\mathrm{mem}}\right)\right),\; \mathbf{U}_{\mathrm{scene}} = \boldsymbol{\gamma}_t \odot \mathbf{U}_{\mathrm{curr}} + (1-\boldsymbol{\gamma}_t)\odot \mathbf{U}_{\mathrm{mem}}$
where $\odot$ is the element-wise multiplication, $\sigma(\cdot)$ denotes the element-wise sigmoid function, so that $\boldsymbol{\gamma}_t \in [0,1]^c$ acts as a soft gate for balancing current-frame and previous scene context.

\noindent\textbf{Explicit Metric Geometry.}
 Scene context features alone do not encode metric distances between interacting parts explicitly, so we inject local 3D support cues through a geometry token. For notational convenience, we write $\mathbf{X}_{t-1}:=\mathbf{X}^{\mathrm{world}}_{t-1}$ for the previous-frame world-frame pointmap. Let $\mathbf{u}_t^n \in \mathbb{R}^{2}$ denote the current anchor of person $n$. For each detected human, we encode the local geometry around $\mathbf{u}_t^n$ into the prompt space:
\begin{equation}
\mathbf{G}_t =
\left\{
\textcolor{learn_red}{\mathrm{MLP}}
\!\left(
\phi_{\mathrm{geo}}\!\left(\mathbf{X}_{t-1}, \mathbf{u}_t^n\right)
\right)
\right\}_{n=1}^{N_t}
\in \mathbb{R}^{N_t \times c},
\end{equation}
where $\phi_{\mathrm{geo}}(\mathbf{X}_{t-1}, \mathbf{u}_t^n)\in\mathbb{R}^{3}$ pools a local pointmap neighborhood centered at $\mathbf{u}_t^n$. Concretely, we construct a fixed-size square window centered at $\mathbf{u}_t^n$ and apply RoIAlign~\cite{he2017mask} to $\mathbf{X}_{t-1}$.

\noindent\textbf{Temporal Momentum.}
Although realistic 3D contact is temporally coherent, reasoning contact from images often tends to be unstable due to strong challenges such as occlusions and depth-scale ambiguities. To stabilize interaction reasoning over time, we reuse the contact token from the previous frame $\mathbf{C}'_{t-1}$ and project it into the current contact prompt latent space as:
\begin{equation}
\mathbf{M}_t =
\textcolor{learn_red}{\mathrm{MLP}}
\!\left(
\mathrm{Align}\!\left(\mathbf{C}'_{t-1}; N_t\right)
\right)
\in \mathbb{R}^{N_t \times c},
\end{equation}
where $\mathrm{Align}(\cdot;N_t)$ keeps the first $\min(N_{t-1},N_t)$ rows of $\mathbf{C}'_{t-1}$, truncates extra rows when $N_{t-1}>N_t$, and appends zero rows when $N_{t-1}<N_t$.

\noindent\textbf{Fusion.}
We fuse human, semantic, geometric, and temporal cues to obtain the final contact token
$\mathbf{C}_t =
\textcolor{learn_red}{\mathrm{MLP}}
\!\left(
\mathbf{H}_t \oplus \mathbf{U}_{\mathrm{scene}} \oplus \mathbf{G}_t \oplus \mathbf{M}_t
\right)$  as shown in \cref{fig:pipeline} (right).

\subsection{Contact-Guided Latent Refinement}
\label{ssec:latent_refinement}

A contact prompt is only useful if it can influence the final body estimate. Unlike Parallel Readout, which predicts contact as an auxiliary output, 
we use the refined contact prompt $\mathbf{C}_t'$ to update the human latent before final \smplx regression as shown in \cref{fig:pipeline} (bottom). First, the human and the contact prompts $\mathbf{H}_t$ and $\mathbf{C}_t$ are processed together by the frozen 4D decoders (\S~\cref{sec:human_branch_prelim}):
\begin{equation}
[\mathbf{F}_t', \mathbf{z}_t', \mathbf{H}_t', \mathbf{C}_t'], \mathbf{S}_t
=
\textcolor{freeze_blue}{\mathrm{Decoders}}\!\left([\mathbf{F}_t, \mathbf{z}_t, \mathbf{H}_t, \mathbf{C}_t],\, \mathbf{S}_{t-1}\right).
\end{equation}

After interaction with the previous state $\mathbf{S}_{t-1}$, we augment $\mathbf{H}_t'$ and $\mathbf{C}_t'$ with the human prior features $\mathbf{F}_{\mathrm{HMR},t}^{\bu}$ obtained from the Multi-HMR encoder as
\begin{equation}
\tilde{\mathbf{H}}_t = \mathbf{H}_t' \oplus \mathbf{F}_{\mathrm{HMR},t}^{\bu}, \qquad
\tilde{\mathbf{C}}_t = \mathbf{C}_t' \oplus \mathbf{F}_{\mathrm{HMR},t}^{\bu}.
\end{equation}

{The refined contact token} $\tilde{\mathbf{C}}_t$ is decoded to obtain dense per-vertex contact logits $\mathbf{s}_t^v$ and probabilities $\hat{\mathbf{c}}_t^v$ on the SMPL mesh ($V_s = 6{,}890$ vertices):
\begin{equation}
\hat{\mathbf{c}}_t^v = \sigma\!\left(\mathbf{s}_t^v\right), \quad \text{ where }
\mathbf{s}_t^v =
\textcolor{learn_red}{\mathrm{Head}_{\text{contact}}}\!\left(\tilde{\mathbf{C}}_t\right)
\in \mathbb{R}^{N_t \times 6890}.
\end{equation}

The final body mesh $\mathbf{Y}_t$ is regressed from a human latent updated by a residual $\Delta \mathbf{H}_t$ based on $\tilde{\mathbf{C}}_t$: %

\begin{equation}
\mathbf{Y}_t = \textcolor{learn_red}{\mathrm{Head}_{\text{human}}}\!\left(\bar{\mathbf{H}}_t\right),\quad \text{ where} \;\; \bar{\mathbf{H}}_t = \tilde{\mathbf{H}}_t + \Delta \mathbf{H}_t
\;\; 
\text{and}\;\; 
 \Delta \mathbf{H}_t = \textcolor{learn_red}{\mathrm{Head}_{\mathrm{residual}}}\!\left(\tilde{\mathbf{C}}_t\right).
\label{eq:residual_update}
\end{equation}

\label{ssec:training}
\noindent\textbf{Training Objectives.}
The overall objective of \modelname $\mathcal{L}_{\mathrm{total}}$ is the sum of
$\mathcal{L}_{\mathrm{4D}}, \mathcal{L}_{\mathrm{SMPLX}}, \mathcal{L}_{\mathrm{contact}}.$
Here, $\mathcal{L}_{\mathrm{4D}}$ denotes the 4D reconstruction losses for metric pointmaps, camera pose, and appearance consistency inherited from CUT3R~\cite{cut3r}, while $\mathcal{L}_{\mathrm{SMPLX}}$ supervises the human output $\mathbf{Y}_t$ using standard SMPL-X~\cite{SMPLX} parameter, mesh, joint, and reprojection losses~\cite{chen2025human3r,cut3r}.

The contact loss is
$\mathcal{L}_{\mathrm{contact}}
=
\mathcal{L}_{c}^{3D}
+
\lambda_{\mathrm{p}}\mathcal{L}_{p},
$
with dense vertex-level supervision
$\mathcal{L}_{c}^{3D}
=
\mathrm{FocalBCE}\!\left(\mathbf{s}_t^v,\mathbf{c}_{\mathrm{gt}}^v\right).
$
Here, $\mathbf{s}_t^v \in \mathbb{R}^{N_t \times 6890}$ are predicted SMPL~\cite{SMPL} contact logits, $\mathbf{c}_{\mathrm{gt}}^v$ are binary vertex-level contact labels, and $\mathcal{L}_{p}$ is a part-level contact loss following DECO~\cite{tripathi2023deco}. Please refer to \textbf{\supmat} (\cref{ssec:impl})for training details.

\section{Experiments}
\label{sec:exp}

\noindent\textbf{Datasets.} For global human motion estimation, we evaluate on RICH~\cite{huang2022capturing} and EMDB-2~\cite{kaufmann2023emdb} following prior works~\cite{chen2025human3r, JOSH3R, li2026unish, zhao2026onlinehmr}. For local human mesh recovery, we evaluate on SLOPER4D~\cite{dai2023sloper4d} and 3DPW~\cite{3DPW}. For physical plausibility and contact prediction, we evaluate on RICH~\cite{huang2022capturing}, since it provides scene geometry, human motion, and dense body-scene contact annotations in world frame. 

\noindent\textbf{Metrics.} Following prior work~\cite{WHAM,TRAM,chen2025human3r}, for global human motion estimation we report WA-MPJPE (mm), W-MPJPE (mm), and RTE (\%). For local human mesh recovery, we report PVE (mm) and MPJPE (mm). For physical plausibility, we report Collision Ratio (\%), Penetrate (cm), Float (cm), and Pen.~Max (cm) using the grounding protocol adapted from HuMoS~\cite{tripathi2024humos}. For contact prediction, we report Precision, Recall, F1, and geometric error, following prior works~\cite{tripathi2023deco, dwivedi2025interactvlm, huang2022capturing}.

\noindent\textbf{Baselines.} We primarily compare with feed-forward monocular 4D human-scene reconstruction methods, as they share \modelname's setting: jointly predicting humans, scene geometry, and camera motion without per-scene optimisation. In particular, we compare against Human3R~\cite{chen2025human3r}, UniSH~\cite{li2026unish}, and JOSH3R~\cite{JOSH3R}. We also introduce two controlled baselines for fair architectural comparisons. $\textrm{Human3R}^{\dagger}$ keeps the original Human3R architecture but uses the same training protocol as \modelname, in which we fine-tune on BEDLAM~\cite{black2023bedlam} and RICH~\cite{huang2022capturing} datasets. \emph{Parallel Readout} adds an auxiliary branch as output to the same backbone without a dedicated contact token, and does not feed contact features back into human reconstruction. Throughout the experiments, $\textrm{Human3R}^{\ast}$ denotes the released Human3R checkpoint evaluated under the same preprocessing, inference, and evaluation pipeline used for \modelname.

\subsection{Global Human Motion Estimation and Local Human Mesh Recovery}
\label{ssec:hmr}

On \textbf{global human motion estimation} (see \cref{tab:global_hmr}), \modelname\ improves over all Human3R-derived baselines on both RICH~\cite{huang2022capturing} and EMDB-2~\cite{kaufmann2023emdb}. 
$\textrm{Human3R}^{\dagger}$ controls for gains induced by the training protocol, while \emph{Parallel Readout} shows the gains come from feeding contact into reconstruction, not predicting it in parallel.
On EMDB-2, \modelname\ again improves over all Human3R-derived baselines, achieving the best WA-MPJPE, RTE, and PA-MPJPE, and the second-best W-MPJPE overall. These results confirm that \modelname's gains generalize beyond RICH. On W-MPJPE, which captures long-horizon trajectory drift by aligning predictions only using the first two frames, UniSH obtains a lower error on EMDB-2, likely benefiting from its sequence-level $\pi^3$ reconstruction backbone. In contrast, \modelname achieves better local pose accuracy, as reflected by the lower WA-MPJPE, while  operating at fast inference speeds (2.4 fps vs 2.0 fps for UniSH), see \cref{fig:teaser}~(b), in a streaming setting.  
Please refer to \textbf{\supmat} (\cref{ssec:hmr_full}) for comparisons against offline optimization methods~\cite{Ye_2023_CVPR_slamhr, TRAM, JOSH3R} and world-grounded HMR~\cite{TRACE, WHAM, GVHMR, zhao2026onlinehmr} methods. 
In \textbf{\supmat} (\cref{ssec:vits672}), we also report global human motion metrics across different backbones 
to show that 
contact-aware \modelname's gains are agnostic to the choice of 
the underlying backbone.

On \textbf{local human mesh recovery} (see \cref{tab:local_mesh}), \modelname\ gives the best PVE and MPJPE among 
baselines on SLOPER4D~\cite{dai2023sloper4d}. 
On the 3DPW~\cite{3DPW} dataset, \modelname\ obtains the lowest MPJPE and PVE comparable to $\textrm{Human3R}^{\ast}$. Overall, the results show that \modelname\ improves global human motion estimation and body-scene alignment without degrading local HMR.

\begin{table*}[t]
\centering
\caption{\textbf{Global human motion estimation on RICH~\cite{huang2022capturing} and EMDB-2~\cite{kaufmann2023emdb}.}
\textbf{Scene} denotes 3D scene reconstruction, and \textbf{Contact} denotes dense human-scene contact prediction or use. $\textrm{Human3R}^{\ast}$: released checkpoint in our pipeline; $\textrm{Human3R}^{\dagger}$: same architecture with our training protocol.
WA-MPJPE, W-MPJPE, and PA-MPJPE are reported in mm; RTE is reported in \%. \textbf{Bold} = best within feed-forward 4D human-scene reconstruction methods; \underline{underlined} = second-best.}
\label{tab:global_hmr}

\footnotesize
\tabstyle
\setlength{\tabcolsep}{2pt}
\begin{adjustbox}{width=\textwidth}
\begin{tabular}{@{}lcccccccccc@{}}
\toprule
\multirow{2}{*}{\textbf{Method}} &
\multirow{2}{*}{\textbf{Scene}} &
\multirow{2}{*}{\textbf{Contact}} &
\multicolumn{4}{c}{\textbf{RICH}~\cite{huang2022capturing}} &
\multicolumn{4}{c}{\textbf{EMDB-2}~\cite{kaufmann2023emdb}} \\
\cmidrule(lr){4-7} \cmidrule(lr){8-11}
& & &
\textbf{WA-MPJPE} $\downarrow$ &
\textbf{W-MPJPE} $\downarrow$ &
\textbf{RTE (\%)} $\downarrow$ &
\textbf{PA-MPJPE} $\downarrow$ &
\textbf{WA-MPJPE} $\downarrow$ &
\textbf{W-MPJPE} $\downarrow$ &
\textbf{RTE (\%)} $\downarrow$ &
\textbf{PA-MPJPE} $\downarrow$ \\
\midrule

JOSH3R~\cite{JOSH3R} & \cmark & \xmark & 190.4 & 334.9 & 6.3 & -- & 220.0 & 661.7 & 13.1 & -- \\
UniSH~\cite{li2026unish} & \cmark & \xmark & 118.1 & 183.2 & 4.8 & 50.0 & 118.5 & \textbf{270.1} & 5.8 & 44.4 \\
$\textrm{Human3R}^{\ast}$~\cite{chen2025human3r} & \cmark & \xmark & 109.9 & 184.0 & 3.3 & 48.6 & \underline{118.2} & 286.1 & \underline{2.4} & \underline{38.7} \\
$\textrm{Human3R}^{\dagger}$~\cite{chen2025human3r} & \cmark & \xmark & \underline{97.5} & \underline{153.2} & \underline{2.3} & \underline{41.4} & -- & -- & -- & -- \\
\addlinespace[2pt]
\cellcolor{gray!10}\textbf{Parallel Readout (Ours)} & \cellcolor{gray!10} \cmark & \cellcolor{gray!10} \cmark & \cellcolor{gray!10} 97.9 & \cellcolor{gray!10} 153.5 & \cellcolor{gray!10} \underline{2.3} & \cellcolor{gray!10} 41.5 & \cellcolor{gray!10} 160.7 & \cellcolor{gray!10} 388.3 & \cellcolor{gray!10} 3.6 & \cellcolor{gray!10} 50.9 \\
\cellcolor{blue!5}\textbf{\modelname\ (Ours)} & \cellcolor{blue!5} \cmark & \cellcolor{blue!5} \cmark & \cellcolor{blue!5} \textbf{81.5} & \cellcolor{blue!5} \textbf{129.8} & \cellcolor{blue!5} \textbf{1.7} & \cellcolor{blue!5} \textbf{37.6} & \cellcolor{blue!5} \textbf{113.7} & \cellcolor{blue!5} \underline{285.6} & \cellcolor{blue!5} \textbf{2.3} & \cellcolor{blue!5} \textbf{38.2} \\
\bottomrule
\end{tabular}
\end{adjustbox}
\end{table*}

\begin{figure*}[t!]
    \centering
    \includegraphics[width=1.0\linewidth]{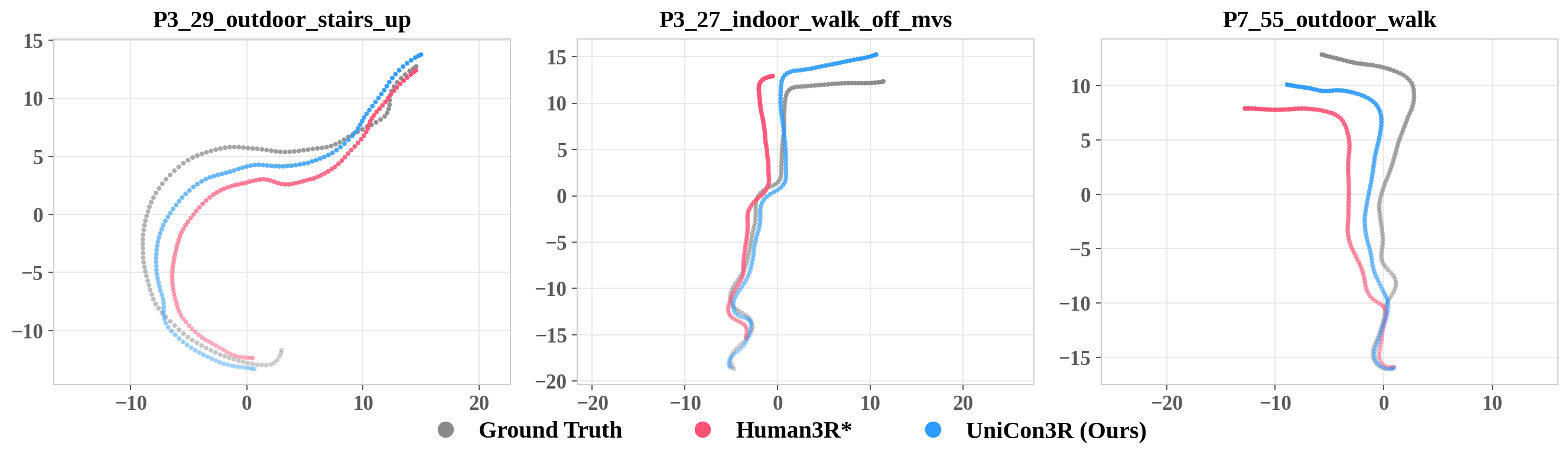}
provide    \caption{\textbf{Comparison of global human trajectory estimation on EMDB-2~\cite{kaufmann2023emdb}.} We compare $\textrm{Human3R}^{\ast}$ and \modelname\ against ground truth after world-coordinate alignment.}
    \label{fig:emdb_comparison}
\end{figure*}

\begin{figure*}[t]
    \centering
    \includegraphics[width=1.0 \linewidth]{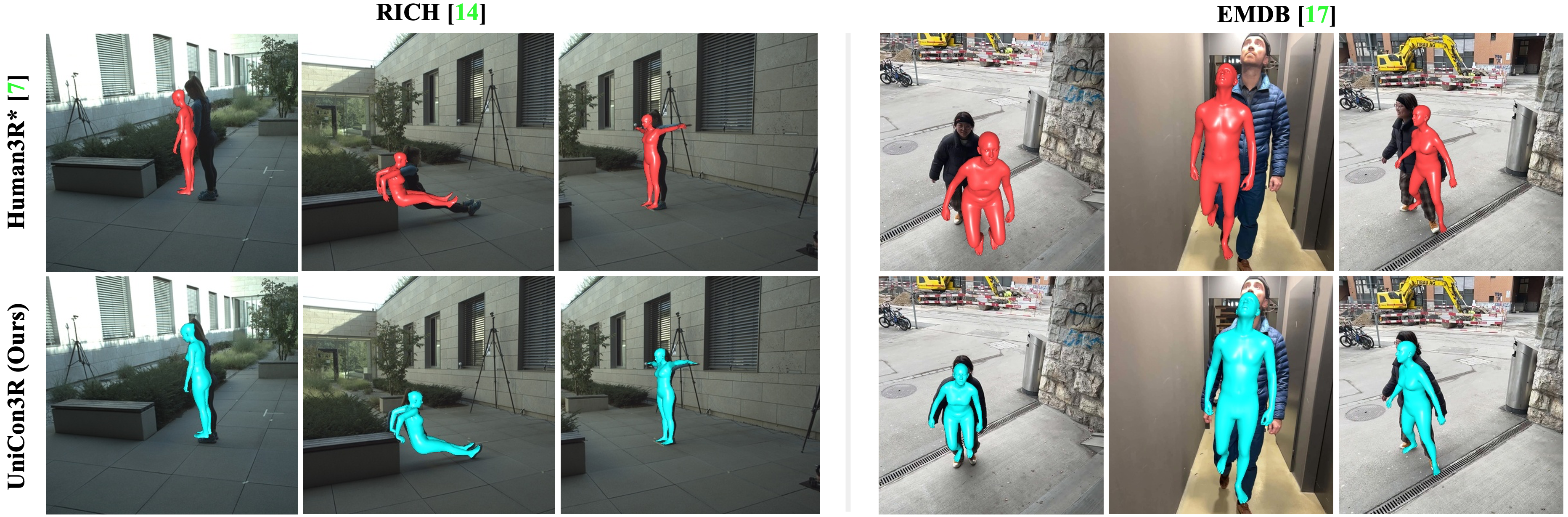}
    \caption{\textbf{Qualitative comparison of local human mesh recovery.} We compare $\textrm{Human3R}^{\ast}$~\cite{chen2025human3r} and \modelname\ against ground truth. \modelname\ produces body pose and scene alignment that are closer to ground truth, particularly in cases with clear scene support.}
    \label{fig:local_mesh_qualitative}
\end{figure*}

The qualitative results in \cref{fig:emdb_comparison,fig:local_mesh_qualitative} show the same pattern. On EMDB-2, \modelname\ follows the ground-truth trajectory more closely than $\textrm{Human3R}^{\ast}$. On SLOPER4D, it produces body pose and scene alignment that are closer to ground truth, especially in cases with clear scene support.
We include additional qualitative results in \supmat video. 

\begin{table}[t]
\centering
\caption{\textbf{Local human mesh reconstruction on SLOPER4D~\cite{dai2023sloper4d} and 3DPW~\cite{3DPW}.}
}
\label{tab:local_mesh}

\scriptsize
\setlength{\tabcolsep}{2pt}
\renewcommand{\arraystretch}{1.0}
\begin{adjustbox}{width=0.8\linewidth}
\begin{tabular}{@{}lccccc@{}}
\toprule
\multirow{2}{*}{\textbf{Method}} &
\multicolumn{3}{c}{\textbf{SLOPER4D}~\cite{dai2023sloper4d}} &
\multicolumn{2}{c}{\textbf{3DPW}~\cite{3DPW}} \\
\cmidrule(lr){2-4} \cmidrule(lr){5-6}
&
\textbf{PVE (mm)} $\downarrow$ &
\textbf{MPJPE (mm)} $\downarrow$ &
\textbf{Pen. Max (cm)} $\downarrow$ &
\textbf{PVE (mm)} $\downarrow$ &
\textbf{MPJPE (mm)} $\downarrow$ \\
\midrule

UniSH~\cite{li2026unish}
& 177.0 & 195.1 & 725.4
& 94.2 & 80.1 \\

$\textrm{Human3R}^{\ast}$~\cite{chen2025human3r}
& \underline{155.1} & \underline{179.1} & \underline{219.7}
& \textbf{86.6} & \underline{72.8} \\

\addlinespace[2pt]
\cellcolor{blue!5}\textbf{\modelname\ (Ours)}
& \cellcolor{blue!5}\textbf{152.9}
& \cellcolor{blue!5}\textbf{176.1}
& \cellcolor{blue!5}\textbf{120.3}
& \cellcolor{blue!5}\underline{86.8}
& \cellcolor{blue!5}\textbf{72.5} \\

\bottomrule
\end{tabular}
\end{adjustbox}
\vspace{-0.5em}
\end{table}

\subsection{Physical Plausibility: Penetration and Floating Meshes}
\label{ssec:phys_eval}

We next evaluate whether the reconstructed body is physically grounded in the recovered scene. \modelname\ reduces \textbf{Pen.~Max} substantially on SLOPER4D~\cite{dai2023sloper4d} as shown in \cref{tab:local_mesh}. \modelname\ outperforms baselines across all metrics also on RICH  as shown in \cref{tab:contact_grounding} (right). Compared with $\textrm{Human3R}^{\ast}$, it reduces \textbf{Penetrate} 
by $\sim 75\%$, \textbf{Float} by $\sim 83\%$
, and \textbf{Pen.~Max} by $\sim 92\%$
. Similar improvements exist over the Parallel Readout baseline. These results show that \modelname\ improves body grounding in world coordinates by reducing scene penetration and unsupported bodies. Note that we do not report Pen.~Max on 3DPW~\cite{3DPW} as the metric assumes a reliable horizontal ground plane, which is often violated in 3DPW, especially in sequences with non-planar or varying support surfaces.

Visualisations in \textbf{\supmat} (\cref{ssec:phys_eval_fig}) show the same trend. Compared with $\textrm{Human3R}^{\ast}$, \modelname\ places the body closer to the support surface and produces more plausible body-scene interaction.

\subsection{Contact Evaluation}
\label{ssec:contact_eval}

\begin{figure*}[t]
    \centering
    \includegraphics[width=1.0\linewidth]{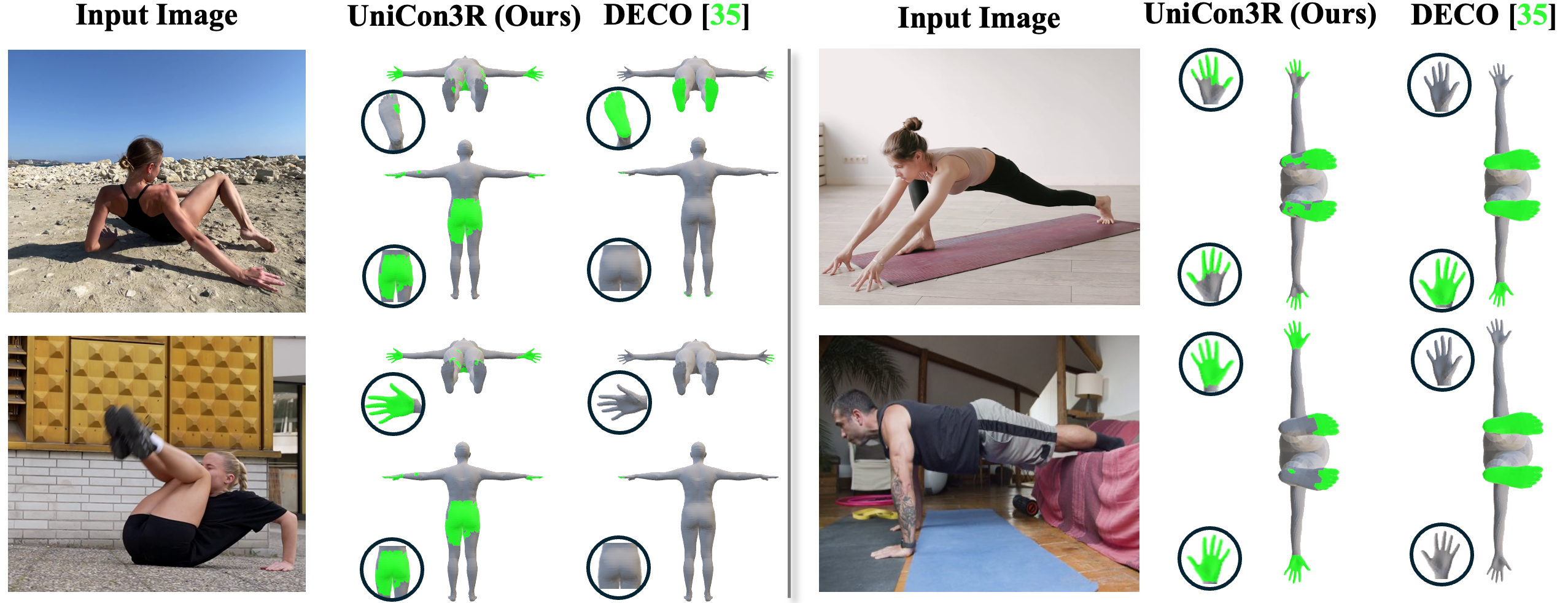}
    \caption{\textbf{Qualitative comparison of contact prediction on web images.} \modelname captures contact on challenging body regions — such as the lower torso, palms, and fingertips — that are partially occluded or lack direct visual evidence, whereas DECO~\cite{tripathi2023deco} tends to concentrate predictions on narrower, visually salient support regions such as the feet.}
    \label{fig:contact_comparison}
\end{figure*}

We evaluate contact prediction on RICH~\cite{huang2022capturing} %
as it 
provides dense body-scene contact annotations together with video sequences, scene geometry, and human motion.
As shown in \cref{tab:contact_grounding} (left), \modelname\ achieves the best F1 score and geodesic error among the compared methods. Compared with \emph{Parallel Readout}, it improves 
F1 by $\sim 14.5\%$, while reducing geodesic error by $\sim 27.5\%$.
Since both models use the same contact supervision, this shows that the gains come from our contact-aware architecture rather than supervision alone. $\textrm{Human3R}^{\ast}$ does not explicitly predict contact, so we derive contact geometrically at test time by thresholding the distance between each predicted human-mesh vertex and its nearest reconstructed scene point at 0.05 m. We also report DECO~\cite{tripathi2023deco} and InteractVLM~\cite{dwivedi2025interactvlm} as off-the-shelf dense-contact estimation baselines. Note that both DECO and InteractVLM are single-frame contact estimation methods, whereas \modelname estimates contact in the challenging video-streaming setting.
While DECO achieves better Precision, \modelname obtains the best Recall, F1, and geodesic error. InteractVLM obtains a geodesic error close to \modelname but is slower at inference, as discussed in \textbf{\supmat}. Qualitatively, \cref{fig:contact_comparison} compares contact prediction performance of \modelname against DECO on arbitrary internet images. DECO often focuses on feet-ground contact, whereas \modelname\ recovers broader body-scene contact patterns. 

In summary, our contact evaluation supports two conclusions. First, contact supervision alone is not sufficient: feeding contact back into the reconstruction pathway improves dense contact prediction over \emph{Parallel Readout}. Second, \modelname\ provides competitive dense contact estimates within a unified reconstruction pipeline, making it 2.0x faster than DECO~\cite{tripathi2023deco} (See \textbf{\supmat}~\cref{ssec:efficiency}).
\begin{table*}[t]
\centering
\caption{\textbf{Contact estimation and physical grounding on RICH~\cite{huang2022capturing}.}
\textbf{Left:} Binary SMPL-vertex contact prediction using precision, recall, F1, and geometric contact error (Geo., cm). \textbf{Right:} physical grounding on reconstructed RICH sequences. Following HuMoS~\cite{tripathi2024humos}, we estimate a robust ground height and use a 5\,mm tolerance to report Collision Ratio (Coll., \%), Penetrate (Pen., cm), Float (cm), and Pen.~Max (P.Max, cm).}
\label{tab:contact_grounding}

\begin{minipage}[t]{0.45\textwidth}
\vspace{0pt}
\centering
\scriptsize
\tabstyle
\renewcommand{\arraystretch}{1.0}
\setlength{\tabcolsep}{3pt}
\begin{tabular}{@{}lcccc@{}}
\toprule
\multirow{2}{*}{\textbf{Method}}
& \multicolumn{4}{c}{\textbf{RICH}~\cite{huang2022capturing}} \\
\cmidrule(lr){2-5}
& \textbf{Prec.} $\uparrow$
& \textbf{Rec.} $\uparrow$
& \textbf{F1} $\uparrow$
& \textbf{Geo.} $\downarrow$ \\
\midrule

\DECO~\cite{tripathi2023deco} & \textbf{0.71} & \underline{0.76} & \underline{0.70} & 17.92 \\
\IVLM~\cite{dwivedi2025interactvlm} & 0.63 & 0.63 & 0.60 & \underline{15.70} \\
$\textrm{Human3R}^{\ast}$~\cite{chen2025human3r} & 0.07 & 0.13 & 0.08 & 76.10 \\
\addlinespace[2pt]
\rowcolor{gray!10} \textbf{Parallel Readout (Ours)} & 0.59 & 0.72 & 0.62 & 20.68 \\
\rowcolor{blue!5} \textbf{\modelname\ (Ours)} & \underline{0.64} & \textbf{0.80} & \textbf{0.71} & \textbf{14.98} \\

\bottomrule
\end{tabular}
\end{minipage}\hfill
\begin{minipage}[t]{0.5\textwidth}
\vspace{0pt}
\centering
\scriptsize
\tabstyle
\renewcommand{\arraystretch}{1.0}
\setlength{\tabcolsep}{3pt}
\begin{tabular}{@{}lcccc@{}}
\toprule
\multirow{2}{*}{\textbf{Method}}
& \multicolumn{4}{c}{\textbf{RICH}~\cite{huang2022capturing}} \\
\cmidrule(lr){2-5}
& \textbf{Coll.} $\downarrow$
& \textbf{Pen.} $\downarrow$
& \textbf{Float} $\downarrow$
& \textbf{P.Max} $\downarrow$ \\
\midrule

UniSH~\cite{li2026unish} & \underline{7.39} & \underline{6.01} & 37.94 & \underline{67.87} \\
$\textrm{Human3R}^{\ast}$~\cite{chen2025human3r} & 7.40 & 6.40 & \underline{33.56} & 216.97 \\
\addlinespace[2pt]
\rowcolor{gray!10} \textbf{Parallel Readout (Ours)} & 7.71 & 7.56 & 37.05 & 254.91 \\
\rowcolor{blue!5} \textbf{\modelname\ (Ours)} & \textbf{5.54} & \textbf{1.59} & \textbf{5.50} & \textbf{17.54} \\

\bottomrule
\end{tabular}
\end{minipage}
\end{table*}

\subsection{Ablation Studies}
\label{ssec:ablation}

We perform a progressive ablation on RICH~\cite{huang2022capturing} to evaluate the contribution of each component in \modelname. Starting from the \emph{Parallel Readout} baseline, we incrementally add scene context, explicit metric geometry, temporal momentum, and contact-guided latent refinement. We report WA-MPJPE and W-MPJPE for global human motion estimation, and Foot Sliding and Jitter to assess grounding stability over time. Results are summarized in \cref{tab:ablation}.

The table also includes $\textrm{Human3R}^{\dagger}$~\cite{chen2025human3r} for reference. Compared with $\textrm{Human3R}^{\dagger}$, Parallel Readout improves Foot Sliding and Jitter, but slightly worsens WA-MPJPE and W-MPJPE. This shows that contact supervision alone can improve temporal grounding metrics, but is not sufficient to improve global motion estimation.
Adding scene context yields the largest single-step improvement over \emph{Parallel Readout}, reducing WA-MPJPE from 97.9 to 89.4~mm and W-MPJPE from 153.5 to 140.6~mm, which highlights the importance of scene-grounded cues for contact reasoning. Adding explicit metric geometry contributes complementary gains, most notably on Foot Sliding (35.0 $\rightarrow$ 33.1~mm), confirming that local 3D support cues further refine grounding. Adding the temporal momentum further improves Foot Sliding and Jitter, indicating better temporal stability. The full model, with contact-guided latent refinement, gives the best result on all four metrics. Relative to \emph{Parallel Readout}, the contact prompt components alone reduce WA-MPJPE from 97.9 to 85.3~mm; adding latent refinement yields a further reduction to 81.5~mm, extending the cumulative gain by 30\% and delivering the best result across all metrics. This shows that predicting contact is not enough: the main improvement comes from feeding contact back into the reconstruction pathway. In other words, contact is most useful when it is corrective rather than descriptive.

\begin{table*}[t]
\centering
\caption{\textbf{Ablation study on RICH~\cite{huang2022capturing}.}
We report global motion and physical plausibility metrics while progressively adding each \modelname\ component. WA-MPJPE and W-MPJPE are in mm. Foot sliding is in mm and Jitter in m/$s^3$, respectively.}
\label{tab:ablation}

\footnotesize
\tabstyle
\setlength{\tabcolsep}{4pt}
\begin{adjustbox}{width=\textwidth}
\begin{tabular}{@{}lcccc|cc|cc@{}}
\toprule
\multirow{2}{*}{\textbf{Model Variant}} &
\multicolumn{4}{c|}{\textbf{Architectural Components}} &
\multicolumn{2}{c|}{\textbf{Global Motion}} &
\multicolumn{2}{c}{\textbf{Physical Plausibility}} \\
\cmidrule(lr){2-5} \cmidrule(lr){6-7} \cmidrule(lr){8-9}
& \textbf{Context} $\mathbf{U}_{\mathrm{scene}}$ & \textbf{Geo.} $\mathbf{G}_t$ & \textbf{Mom.} $\mathbf{M}_t$ & \textbf{Latent Ref.}
& \textbf{WA-MPJPE} $\downarrow$ & \textbf{W-MPJPE} $\downarrow$
& \textbf{Foot Sliding} $\downarrow$ & \textbf{Jitter} $\downarrow$ \\
\midrule

$\textrm{Human3R}^{\dagger}$~\cite{chen2025human3r} & -- & -- & -- & -- & 97.5 & 153.2 & 37.7 & 279.5 \\
\rowcolor{gray!10} Parallel Readout & \xmark & \xmark & \xmark & \xmark & 97.9 & 153.5 & 35.3 & 262.5 \\
\rowcolor{gray!10} + Scene Context & \cmark & \xmark & \xmark & \xmark & 89.4 & 140.6 & 35.0 & 260.1 \\
\rowcolor{gray!10} + Explicit Geometry & \cmark & \cmark & \xmark & \xmark & 85.9 & \underline{134.5} & 33.1 & 244.6 \\
\rowcolor{gray!10} + Temporal Momentum & \cmark & \cmark & \cmark & \xmark & \underline{85.3} & 135.1 & \underline{32.0} & \underline{231.8} \\
\midrule
\addlinespace[2pt]
\rowcolor{blue!5} \textbf{+ Latent Refinement (\modelname)} & \cmark & \cmark & \cmark & \cmark & \textbf{81.5} & \textbf{129.8} & \textbf{31.5} & \textbf{221.4} \\
\bottomrule
\end{tabular}
\end{adjustbox}
\end{table*}

\vspace{-2mm}

\subsection{\modelname\ with Test-Time Optimization}
\label{ssec:tto}

\modelname\ is designed for online feed-forward inference, and all main comparisons use this setting. 
Fig.~\ref{fig:teaser}(b) additionally reports an optional post-hoc variant, \modelname{}$^{\ast}$, which applies a JOSH-style~\cite{JOSH3R} camera refinement while keeping the \modelname\ body prediction fixed. 
This variant is not part of the default pipeline; it only illustrates the accuracy--runtime trade-off when per-sequence refinement is allowed. 
\modelname{}$^{\ast}$ applies a JOSH-style~\cite{JOSH3R} camera refinement while keeping the \modelname\ body prediction fixed. \modelname\ achieves 81.5\,mm WA-MPJPE$_{100}$ at 2.40 FPS, compared with JOSH (102.6\,mm, 0.16 FPS), UniSH (118.1\,mm, 2.00 FPS), and $\textrm{Human3R}^{\dagger}$ (97.5\,mm, 2.41 FPS); the optional test-time optimized variant \modelname$^{\ast}$ further reduces the error to 72.2\,mm.
\vspace{-2mm}

\section{Conclusion}
\label{sec:conclusion}
We presented \textbf{\modelname}, a unified feed-forward framework for 4D human-scene reconstruction from monocular video that uses contact inside the reconstruction pathway. By introducing a scene-aware contact prompt and contact-guided latent refinement, \modelname\ improves body grounding while preserving the simplicity and efficiency of unified feed-forward inference. 
Across benchmarks, the results support the conclusion that 
in unified feed-forward human-scene reconstruction, predicting contact as an auxiliary output is not sufficient; contact must directly influence the final body estimate. Via the \modelname framework, we show that this paradigm results in complementary gains; better grounding, better world-coordinate motion estimation, and local mesh accuracy, all achievable with a single end-to-end model.

\noindent\textbf{Limitations and Future Directions.}
Several limitations remain. In our work, contact is represented as binary per-vertex labels rather than rich physical quantities such as proximity, contact forces or friction. The quality of recovered contact often depends on the quality of the reconstructed scene geometry, and we observe that scene errors can propagate to contact prediction and body-scene alignment. Future work may explore ways to improve scene reconstruction using contact. Our grounding and plausibility metrics assume rigid support geometry that is approximated by the reconstructed scene; more complex settings such as deformable objects, non-rigid scenes, or fine-grained object interactions are not explicitly modeled. Finally, we observe that augmenting \modelname{} with a post-hoc test-time optimization stage further reduces global trajectory error.

\bibliographystyle{plainnat}
\bibliography{main}

@STRING{CVPR = {Proc. IEEE Conf. on Computer Vision and Pattern	Recognition (CVPR)}}

@STRING{ECCV = {Proc. of the European Conf. on Computer Vision (ECCV)}}

@STRING{ICCV = {Proc. of the IEEE International Conf. on Computer Vision (ICCV)}}

@STRING{SIGGRAPH = {ACM Trans. on Graphics}}

@STRING{THREEDV = {Proc. of the International Conf. on 3D Vision (3DV)}}

@STRING{COMPUTER = {IEEE Computer}}

@STRING{JOS = {Journal of the Optical Society America (JOSA)}}

@STRING{ARXIV = {arXiv.org}}

@inproceedings{GVHMR,
  title={World-grounded human motion recovery via gravity-view coordinates},
  author={Shen, Zehong and Pi, Huaijin and Xia, Yan and Cen, Zhi and Peng, Sida and Hu, Zechen and Bao, Hujun and Hu, Ruizhen and Zhou, Xiaowei},
  booktitle={SIGGRAPH Asia},
  year={2024}
}

@inproceedings{TRAM,
  title={TRAM: Global Trajectory and Motion of 3D Humans from in-the-wild Videos},
  author={Wang, Yufu and Wang, Ziyun and Liu, Lingjie and Daniilidis, Kostas},
  booktitle=ECCV,
  year={2024}
}

@inproceedings{pavlakos2022one,
  title={The one where they reconstructed 3d humans and environments in tv shows},
  author={Pavlakos, Georgios and Weber, Ethan and Tancik, Matthew and Kanazawa, Angjoo},
  booktitle={European Conference on Computer Vision},
  pages={732--749},
  year={2022},
  organization={Springer}
}

@inproceedings{JOSH3R,
  title={Joint optimization for 4d human-scene reconstruction in the wild},
  author={Liu, Zhizheng and Lin, Joe and Wu, Wayne and Zhou, Bolei},
  booktitle={The Fourteenth International Conference on Learning Representations},
  year={2026},
}

@inproceedings{TRACE,
  title={Trace: 5d temporal regression of avatars with dynamic cameras in 3d environments},
  author={Sun, Yu and Bao, Qian and Liu, Wu and Mei, Tao and Black, Michael J},
  booktitle=CVPR,
  year={2023}
}

@inproceedings{WHAM,
  title={Wham: Reconstructing world-grounded humans with accurate 3d motion},
  author={Shin, Soyong and Kim, Juyong and Halilaj, Eni and Black, Michael J},
  booktitle=CVPR,
  year={2024}
}

@inproceedings{CameraHMR,
  title={Camerahmr: Aligning people with perspective},
  author={Patel, Priyanka and Black, Michael J},
  booktitle=THREEDV,
  year={2025}
}

@inproceedings{Multi-HMR,
  title={Multi-hmr: Multi-person whole-body human mesh recovery in a single shot},
  author={Baradel, Fabien and Armando, Matthieu and Galaaoui, Salma and Br{\'e}gier, Romain and Weinzaepfel, Philippe and Rogez, Gr{\'e}gory and Lucas, Thomas},
  booktitle=ECCV,
  year={2024},
}

@inproceedings{vpt,
  title={Visual prompt tuning},
  author={Jia, Menglin and Tang, Luming and Chen, Bor-Chun and Cardie, Claire and Belongie, Serge and Hariharan, Bharath and Lim, Ser-Nam},
  booktitle={European conference on computer vision},
  pages={709--727},
  year={2022},
  organization={Springer}
}

@inproceedings{cut3r,
  author={Qianqian Wang and Yifei Zhang and Aleksander Holynski and Alexei A. Efros and Angjoo Kanazawa},
  title={Continuous 3D Perception Model with Persistent State},
  booktitle=CVPR,
  year={2025}
}

@inproceedings{3DPW,
  title={Recovering accurate 3d human pose in the wild using imus and a moving camera},
  author={Von Marcard, Timo and Henschel, Roberto and Black, Michael J and Rosenhahn, Bodo and Pons-Moll, Gerard},
  booktitle={Proceedings of the European conference on computer vision (ECCV)},
  pages={601--617},
  year={2018}
}

@incollection{SMPL,
  title={SMPL: A skinned multi-person linear model},
  author={Loper, Matthew and Mahmood, Naureen and Romero, Javier and Pons-Moll, Gerard and Black, Michael J},
  booktitle={Seminal Graphics Papers: Pushing the Boundaries, Volume 2},
  pages={851--866},
  year={2023}
}

@inproceedings{SMPLX,
  title={Expressive body capture: 3d hands, face, and body from a single image},
  author={Pavlakos, Georgios and Choutas, Vasileios and Ghorbani, Nima and Bolkart, Timo and Osman, Ahmed AA and Tzionas, Dimitrios and Black, Michael J},
  booktitle={Proceedings of the IEEE/CVF conference on computer vision and pattern recognition},
  pages={10975--10985},
  year={2019}
}

@inproceedings{bogo2016keep,
  title={Keep it SMPL: Automatic estimation of 3D human pose and shape from a single image},
  author={Bogo, Federica and Kanazawa, Angjoo and Lassner, Christoph and Gehler, Peter and Romero, Javier and Black, Michael J},
  booktitle={European conference on computer vision},
  pages={561--578},
  year={2016},
  organization={Springer}
}

@inproceedings{black2023bedlam,
  title={Bedlam: A synthetic dataset of bodies exhibiting detailed lifelike animated motion},
  author={Black, Michael J and Patel, Priyanka and Tesch, Joachim and Yang, Jinlong},
  booktitle={Proceedings of the IEEE/CVF Conference on Computer Vision and Pattern Recognition},
  pages={8726--8737},
  year={2023}
}

@inproceedings{dwivedi2024tokenhmr,
  title={Tokenhmr: Advancing human mesh recovery with a tokenized pose representation},
  author={Dwivedi, Sai Kumar and Sun, Yu and Patel, Priyanka and Feng, Yao and Black, Michael J},
  booktitle={Proceedings of the IEEE/CVF conference on computer vision and pattern recognition},
  pages={1323--1333},
  year={2024}
}

@inproceedings{goel2023humans,
  title={Humans in 4d: Reconstructing and tracking humans with transformers},
  author={Goel, Shubham and Pavlakos, Georgios and Rajasegaran, Jathushan and Kanazawa, Angjoo and Malik, Jitendra},
  booktitle={Proceedings of the IEEE/CVF International Conference on Computer Vision},
  pages={14783--14794},
  year={2023}
}

@inproceedings{wang2023refit,
  title={Refit: Recurrent fitting network for 3d human recovery},
  author={Wang, Yufu and Daniilidis, Kostas},
  booktitle={Proceedings of the IEEE/CVF International Conference on Computer Vision},
  pages={14644--14654},
  year={2023}
}

@inproceedings{kanazawa2018end,
  title={End-to-end recovery of human shape and pose},
  author={Kanazawa, Angjoo and Black, Michael J and Jacobs, David W and Malik, Jitendra},
  booktitle={Proceedings of the IEEE conference on computer vision and pattern recognition},
  pages={7122--7131},
  year={2018}
}

@inproceedings{li2022cliff,
  title={Cliff: Carrying location information in full frames into human pose and shape estimation},
  author={Li, Zhihao and Liu, Jianzhuang and Zhang, Zhensong and Xu, Songcen and Yan, Youliang},
  booktitle={European Conference on Computer Vision},
  pages={590--606},
  year={2022},
  organization={Springer}
}

@article{xu2023smpler,
  title={Smpler: Taming transformers for monocular 3d human shape and pose estimation},
  author={Xu, Xiangyu and Liu, Lijuan and Yan, Shuicheng},
  journal={IEEE Transactions on Pattern Analysis and Machine Intelligence},
  volume={46},
  number={5},
  pages={3275--3289},
  year={2023},
  publisher={IEEE}
}

@inproceedings{cai2023smpler,
  title={{SMPL}er-{X}: Scaling up expressive human pose and shape estimation},
  author={Cai, Zhongang and Yin, Wanqi and Zeng, Ailing and Wei, Chen and Sun, Qingping and Yanjun, Wang and Pang, Hui En and Mei, Haiyi and Zhang, Mingyuan and Zhang, Lei and others},
  journal={Advances in Neural Information Processing Systems},
  volume={36},
  pages={11454--11468},
  year={2023}
}

@article{yin2025smplest,
  title={Smplest-x: Ultimate scaling for expressive human pose and shape estimation},
  author={Yin, Wanqi and Cai, Zhongang and Wang, Ruisi and Zeng, Ailing and Wei, Chen and Sun, Qingping and Mei, Haiyi and Wang, Yanjun and Pang, Hui En and Zhang, Mingyuan and others},
  journal={arXiv preprint arXiv:2501.09782},
  year={2025}
}

@inproceedings{muller2025reconstructing,
  title={Reconstructing people, places, and cameras},
  author={M{\"u}ller, Lea and Choi, Hongsuk and Zhang, Anthony and Yi, Brent and Malik, Jitendra and Kanazawa, Angjoo},
  booktitle={Proceedings of the Computer Vision and Pattern Recognition Conference},
  pages={21948--21958},
  year={2025}
}

@article{rojas2025hamst3r,
  title={HAMSt3R: Human-Aware Multi-view Stereo 3D Reconstruction},
  author={Rojas, Sara and Armando, Matthieu and Ghamen, Bernard and Weinzaepfel, Philippe and Leroy, Vincent and Rogez, Gregory},
  journal={arXiv preprint arXiv:2508.16433},
  year={2025}
}

@inproceedings{chen2025human3r,
  title={Human3R: Everyone Everywhere All at Once},
  author={Chen, Yue and Chen, Xingyu and Xue, Yuxuan and Chen, Anpei and Xiu, Yuliang and Pons-Moll, Gerard},
  booktitle={The Fourteenth International Conference on Learning Representations},
  year={2026},
}

@inproceedings{zhao2026onlinehmr,
  title={OnlineHMR: Video-based Online World-Grounded Human Mesh Recovery},
  author={Zhao, Yiwen and Zheng, Ce and Wang, Yufu and Yang, Hsueh-Han Daniel and Wen, Liting and Jeni, Laszlo A.},
  booktitle={CVPR},
  year={2026}
}

@inproceedings{kaufmann2023emdb,
  author = {Kaufmann, Manuel and Song, Jie and Guo, Chen and Shen, Kaiyue and Jiang, Tianjian and Tang, Chengcheng and Z{\'a}rate, Juan Jos{\'e} and Hilliges, Otmar},
  title = {{EMDB}: The {E}lectromagnetic {D}atabase of {G}lobal 3{D} {H}uman {P}ose and {S}hape in the {W}ild},
  booktitle = {International Conference on Computer Vision (ICCV)},
  year = {2023}
}

@InProceedings{Ye_2023_CVPR_slamhr,
    author    = {Ye, Vickie and Pavlakos, Georgios and Malik, Jitendra and Kanazawa, Angjoo},
    title     = {Decoupling Human and Camera Motion From Videos in the Wild},
    booktitle = {Proceedings of the IEEE/CVF Conference on Computer Vision and Pattern Recognition (CVPR)},
    month     = {June},
    year      = {2023},
    pages     = {21222-21232}
}

@inproceedings{zhao2024synergistic,
  title={Synergistic global-space camera and human reconstruction from videos},
  author={Zhao, Yizhou and Wang, Tuanfeng Yang and Raj, Bhiksha and Xu, Min and Yang, Jimei and Huang, Chun-Hao Paul},
  booktitle={Proceedings of the IEEE/CVF Conference on Computer Vision and Pattern Recognition},
  pages={1216--1226},
  year={2024}
}

@inproceedings{dwivedi2025interactvlm,
  title={InteractVLM: 3D interaction reasoning from 2D foundational models},
  author={Dwivedi, Sai Kumar and Anti{\'c}, Dimitrije and Tripathi, Shashank and Taheri, Omid and Schmid, Cordelia and Black, Michael J and Tzionas, Dimitrios},
  booktitle={Proceedings of the Computer Vision and Pattern Recognition Conference},
  pages={22605--22615},
  year={2025}
}

@inproceedings{huang2022capturing,
  title={Capturing and inferring dense full-body human-scene contact},
  author={Huang, Chun-Hao P and Yi, Hongwei and H{\"o}schle, Markus and Safroshkin, Matvey and Alexiadis, Tsvetelina and Polikovsky, Senya and Scharstein, Daniel and Black, Michael J},
  booktitle={Proceedings of the IEEE/CVF Conference on Computer Vision and Pattern Recognition},
  pages={13274--13285},
  year={2022}
}

@inproceedings{tripathi2023deco,
  title={Deco: Dense estimation of 3d human-scene contact in the wild},
  author={Tripathi, Shashank and Chatterjee, Agniv and Passy, Jean-Claude and Yi, Hongwei and Tzionas, Dimitrios and Black, Michael J},
  booktitle={Proceedings of the IEEE/CVF international conference on computer vision},
  pages={8001--8013},
  year={2023}
}

@inproceedings{li2026unish,
    title={UniSH: Unifying Scene and Human Reconstruction in a Feed-Forward Pass},
  author={Li, Mengfei and Li, Peng and Zhang, Zheng and Lu, Jiahao and Zhao, Chengfeng and Xue, Wei and Liu, Qifeng and Peng, Sida and Zhang, Wenxiao and Luo, Wenhan and others},
    booktitle = {Conference on Computer Vision and Pattern Recognition ({CVPR})},
    year = {2026},
}

@inproceedings{yalandur2025physic,
  title={Physic: Physically plausible 3d human-scene interaction and contact from a single image},
  author={Yalandur Muralidhar, Pradyumna and Xue, Yuxuan and Xie, Xianghui and Kostyrko, Margaret and Pons-Moll, Gerard},
  booktitle={Proceedings of the SIGGRAPH Asia 2025 Conference Papers},
  pages={1--12},
  year={2025}
}

@inproceedings{weng2021holistic,
  title={Holistic 3d human and scene mesh estimation from single view images},
  author={Weng, Zhenzhen and Yeung, Serena},
  booktitle={Proceedings of the IEEE/CVF Conference on Computer Vision and Pattern Recognition},
  pages={334--343},
  year={2021}
}

@inproceedings{tripathi2024humos,
  title={Humos: Human motion model conditioned on body shape},
  author={Tripathi, Shashank and Taheri, Omid and Lassner, Christoph and Black, Michael and Holden, Daniel and Stoll, Carsten},
  booktitle={European Conference on Computer Vision},
  pages={133--152},
  year={2024},
  organization={Springer}
}

@inproceedings{kocabas2024pace,
  title={Pace: Human and camera motion estimation from in-the-wild videos},
  author={Kocabas, Muhammed and Yuan, Ye and Molchanov, Pavlo and Guo, Yunrong and Black, Michael J and Hilliges, Otmar and Kautz, Jan and Iqbal, Umar},
  booktitle={2024 International Conference on 3D Vision (3DV)},
  pages={397--408},
  year={2024},
  organization={IEEE}
}

@inproceedings{li2024coin,
  title={Coin: Control-inpainting diffusion prior for human and camera motion estimation},
  author={Li, Jiefeng and Yuan, Ye and Rempe, Davis and Zhang, Haotian and Molchanov, Pavlo and Lu, Cewu and Kautz, Jan and Iqbal, Umar},
  booktitle={European Conference on Computer Vision},
  pages={426--446},
  year={2024},
  organization={Springer}
}

@inproceedings{sun2022putting,
  title={Putting people in their place: Monocular regression of 3d people in depth},
  author={Sun, Yu and Liu, Wu and Bao, Qian and Fu, Yili and Mei, Tao and Black, Michael J},
  booktitle={Proceedings of the IEEE/CVF Conference on Computer Vision and Pattern Recognition},
  pages={13243--13252},
  year={2022}
}

@inproceedings{villegas2021contact,
  title={Contact-aware retargeting of skinned motion},
  author={Villegas, Ruben and Ceylan, Duygu and Hertzmann, Aaron and Yang, Jimei and Saito, Jun},
  booktitle={Proceedings of the IEEE/CVF International Conference on Computer Vision},
  pages={9720--9729},
  year={2021}
}

@inproceedings{cseke2025pico,
  title={PICO: Reconstructing 3D people in contact with objects},
  author={Cseke, Alp{\'a}r and Tripathi, Shashank and Dwivedi, Sai Kumar and Lakshmipathy, Arjun S and Chatterjee, Agniv and Black, Michael J and Tzionas, Dimitrios},
  booktitle={Proceedings of the Computer Vision and Pattern Recognition Conference},
  pages={1783--1794},
  year={2025}
}

@inproceedings{xue2024hsr,
  title={HSR: holistic 3d human-scene reconstruction from monocular videos},
  author={Xue, Lixin and Guo, Chen and Zheng, Chengwei and Wang, Fangjinghua and Jiang, Tianjian and Ho, Hsuan-I and Kaufmann, Manuel and Song, Jie and Hilliges, Otmar},
  booktitle={European Conference on Computer Vision},
  pages={429--448},
  year={2024},
  organization={Springer}
}

@inproceedings{tripathi2023ipman,
    title = {{3D} Human Pose Estimation via Intuitive Physics},
    author = {Tripathi, Shashank and M{\"u}ller, Lea and Huang, Chun-Hao P. and Taheri Omid and Black, Michael J. and Tzionas, Dimitrios},
    booktitle = {Conference on Computer Vision and Pattern Recognition ({CVPR})},
    pages = {4713--4725},
    year = {2023},
    url = {https://ipman.is.tue.mpg.de}
}

@inproceedings{dai2023sloper4d,
  title={Sloper4d: A scene-aware dataset for global 4d human pose estimation in urban environments},
  author={Dai, Yudi and Lin, YiTai and Lin, XiPing and Wen, Chenglu and Xu, Lan and Yi, Hongwei and Shen, Siqi and Ma, Yuexin and Wang, Cheng},
  booktitle={Proceedings of the IEEE/CVF conference on computer vision and pattern recognition},
  pages={682--692},
  year={2023}
}

@inproceedings{he2017mask,
  title={Mask r-cnn},
  author={He, Kaiming and Gkioxari, Georgia and Doll{\'a}r, Piotr and Girshick, Ross},
  booktitle={Proceedings of the IEEE international conference on computer vision},
  pages={2961--2969},
  year={2017}
}

@inproceedings{voita2019analyzing,
  title={Analyzing Multi-Head Self-Attention: Specialized Heads Do the Heavy Lifting, the Rest Can Be Pruned},
  author={Voita, Elena and Talbot, David and Moiseev, Fedor and Sennrich, Rico and Titov, Ivan},
  booktitle={Proceedings of the 57th Annual Meeting of the Association for Computational Linguistics},
  pages={5797--5808},
  year={2019}
}

@inproceedings{michel2019sixteen,
  title={Are Sixteen Heads Really Better than One?},
  author={Michel, Paul and Levy, Omer and Neubig, Graham},
  booktitle={Advances in Neural Information Processing Systems},
  volume={32},
  year={2019}
}

@inproceedings{abnar2020quantifying,
  title={Quantifying Attention Flow in Transformers},
  author={Abnar, Samira and Zuidema, Willem},
  booktitle={Proceedings of the 58th Annual Meeting of the Association for Computational Linguistics},
  pages={4190--4197},
  year={2020}
}

@inproceedings{vig2020investigating,
  title={Investigating Gender Bias in Language Models Using Causal Mediation Analysis},
  author={Vig, Jesse and Gehrmann, Sebastian and Belinkov, Yonatan and Qian, Sharon and Nevo, Daniel and Singer, Yaron and Shieber, Stuart},
  booktitle={Advances in Neural Information Processing Systems},
  volume={33},
  pages={12388--12401},
  year={2020}
}

@inproceedings{chefer2021transformer,
  title={Transformer Interpretability Beyond Attention Visualization},
  author={Chefer, Hila and Gur, Shir and Wolf, Lior},
  booktitle={Proceedings of the IEEE/CVF Conference on Computer Vision and Pattern Recognition},
  pages={782--791},
  year={2021}
}

@inproceedings{meng2022locating,
  title={Locating and Editing Factual Associations in {GPT}},
  author={Meng, Kevin and Bau, David and Andonian, Alex and Belinkov, Yonatan},
  booktitle={Advances in Neural Information Processing Systems},
  volume={35},
  year={2022}
}

\appendix

\clearpage

\begin{center}
{\Large\bfseries Appendix}
\end{center}

\vspace{1.5em}

{\noindent\Large\bfseries Table of Contents}
\vspace{0.3em}
\hrule
\vspace{0.8em}

\noindent\textbf{A\quad Analysis}\hfill\pageref{app:analysis}\\[0.25em]
\hspace*{1.2em}A.1\quad Empirical Analysis of the Contact Pathway\dotfill\pageref{ssec:empirical_internal}\\
\hspace*{1.2em}A.2\quad Comprehensive Comparison for Global Human Motion Estimation\dotfill\pageref{ssec:hmr_full}\\
\hspace*{1.2em}A.3\quad Runtime Efficiency\dotfill\pageref{ssec:efficiency}\\
\hspace*{1.2em}A.4\quad Backbone Scaling: Lightweight ViT-S/672 Variant\dotfill\pageref{ssec:vits672}\\[0.6em]

\noindent\textbf{B\quad Implementation Details}\hfill\pageref{ssec:impl}\\[0.6em]

\noindent\textbf{C\quad Physical Plausibility: Qualitative Examples}\hfill\pageref{ssec:phys_eval_fig}

\vspace{0.8em}
\hrule
\vspace{2em}

\section{Analysis}
\label{app:analysis}

\subsection{Empirical Analysis of the Contact Pathway}
\label{ssec:empirical_internal}

To complement the downstream reconstruction results, we report a small diagnostic analysis of how contact is used inside \modelname. The goal is not to claim that attention alone explains model behavior, but to test whether the contact pathway measurably changes the internal computation that produces the final body estimate. Following common practice in transformer analysis, we pair attention-based observations with explicit controls and a counterfactual intervention on the latent residual branch~\cite{abnar2020quantifying,chefer2021transformer,voita2019analyzing,michel2019sixteen,vig2020investigating,meng2022locating}.

\paragraph{Evaluation protocol.}
We evaluate on the moving-camera subset of RICH~\cite{huang2022capturing}, which contains $40$ sequences. We compare three checkpoints under identical inputs and instrumentation: $\textrm{Human3R}^{\dagger}$, Parallel Readout, and \modelname. $\textrm{Human3R}^{\dagger}$ keeps the original Human3R architecture and uses the same training protocol as \modelname, while Parallel Readout adds contact supervision without feeding the contact representation back into the human reconstruction pathway. Relative to these baselines, \modelname\ differs in the two contributions of the main paper: it introduces a \emph{scene-aware contact prompt} $\mathbf{C}_t$, constructed from semantic scene context, explicit metric geometry, and a temporal prior, and it applies \emph{contact-guided latent refinement}, which feeds the refined contact representation back into the HMR pathway before final SMPL-X~\cite{SMPLX} regression.

\begin{figure*}[t]
    \centering
    \begin{subfigure}[t]{0.48\linewidth}
        \centering
        \includegraphics[width=\linewidth]{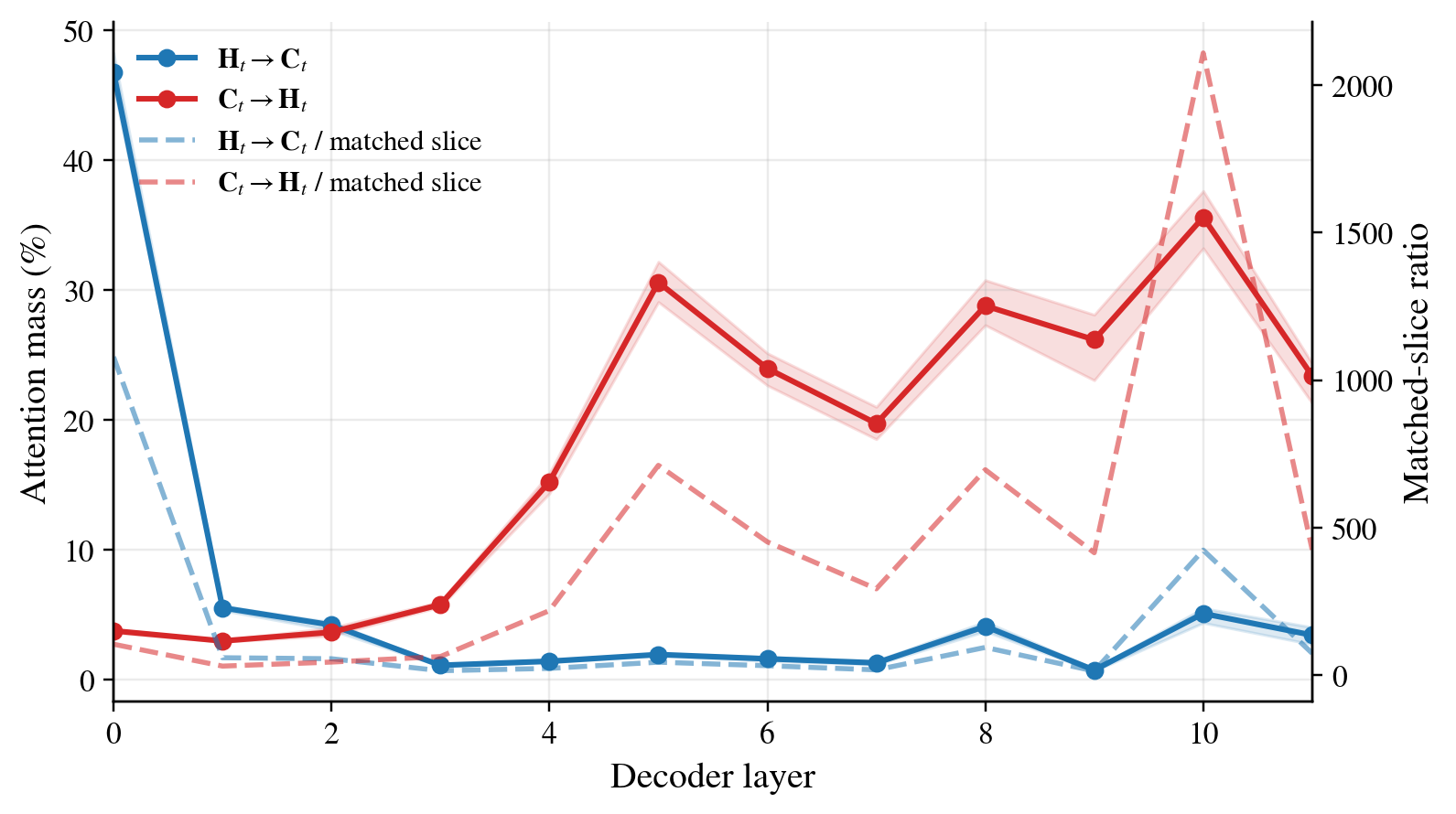}
    \end{subfigure}\hfill
    \begin{subfigure}[t]{0.48\linewidth}
        \centering
        \includegraphics[width=\linewidth]{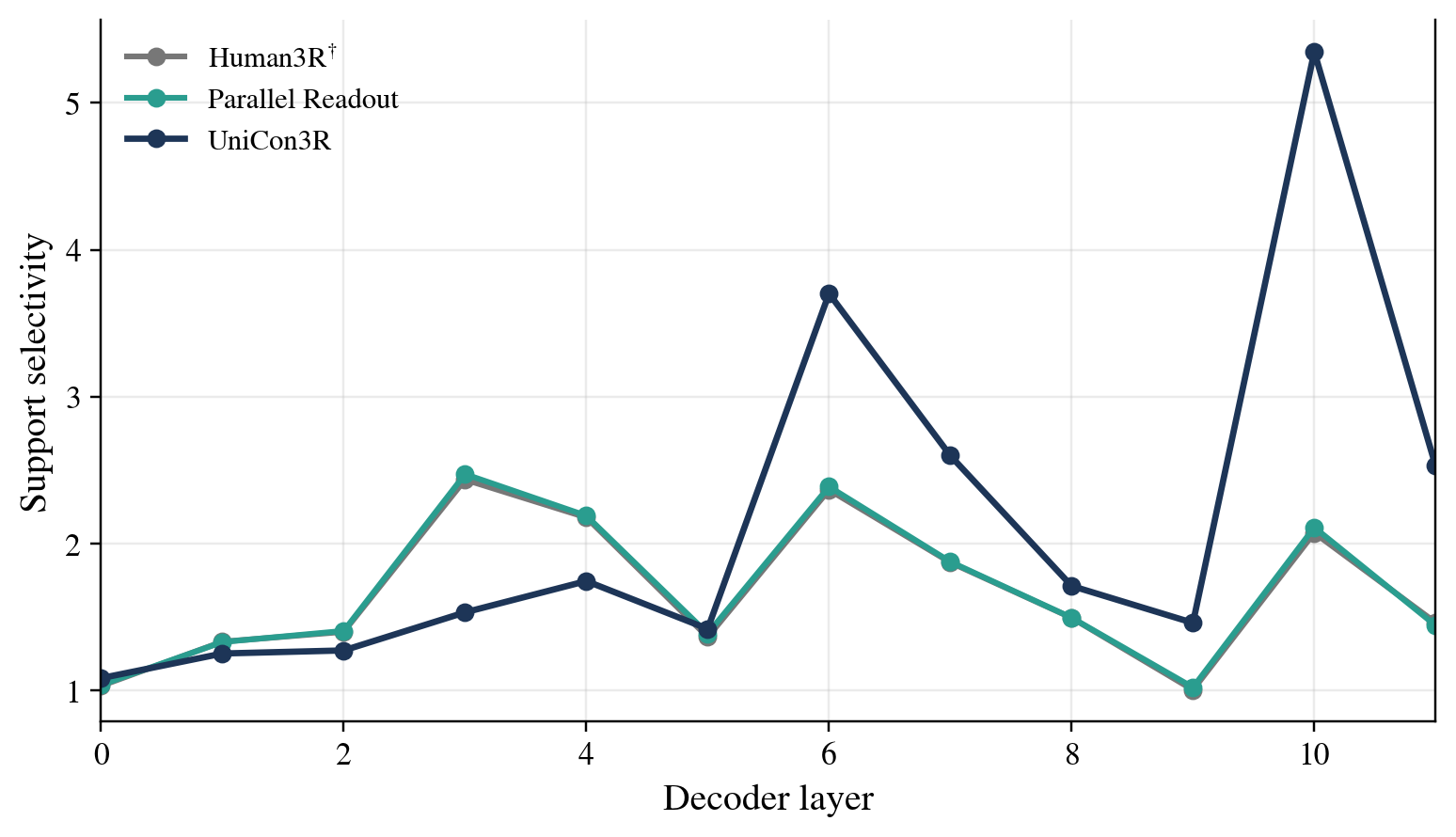}
    \end{subfigure}
    \caption{\textbf{Empirical Analysis of the Contact Pathway.} \textbf{Left:} layerwise human--contact coupling inside the joint decoder token sequence. Solid curves denote the mean coupling mass (left axis), showing strongest early $\mathbf{H}_t\!\rightarrow\!\mathbf{C}_t$ coupling and strongest late $\mathbf{C}_t\!\rightarrow\!\mathbf{H}_t$ coupling. Dashed curves denote the corresponding matched-slice ratios (right axis), where each coupling is normalized by a matched control from a same-size slice of image tokens in $\mathbf{F}_t$. \textbf{Right:} support-region selectivity of the human stream. \modelname\ allocates more attention mass to support-relevant image regions than $\textrm{Human3R}^{\dagger}$ and Parallel Readout. Together, these diagnostics support that contact is used inside the reconstruction pathway as an internal corrective signal.}
    \label{fig:supp_internal_mechanism}
\end{figure*}

\paragraph{Human--contact routing.}
We first analyze layerwise coupling between the human token and the contact token inside the joint decoder token sequence $[\mathbf{z},\mathbf{F}_t,\mathbf{H}_t,\mathbf{C}_t]$ processed by the frozen 4D decoders inherited from Human3R~\cite{chen2025human3r}/CUT3R~\cite{cut3r}. Fig.~\ref{fig:supp_internal_mechanism}\textbf{ (left)} plots the mean coupling in both directions. The goal is not to treat attention weights as explanations by themselves, but to test whether the \emph{scene-aware contact prompt} participates in the same decoder computation as the human prompt. To do so, we compare the human--contact coupling against a matched control derived from ordinary image tokens. Concretely, for each human--contact interaction, we replace the target token by a same-size slice drawn from the image-token set $\mathbf{F}_t$ and measure the corresponding attention mass under the same decoder layer and head. This gives a baseline for how much attention would be assigned to a generic token interaction under otherwise identical conditions. If the human--contact coupling is much larger than this baseline, then the effect is unlikely to be explained by generic attention concentration alone. As observed in Fig.~\ref{fig:supp_internal_mechanism}\textbf{ (left)}, across the moving-camera subset, the strongest $\mathbf{H}_t\!\rightarrow\!\mathbf{C}_t$ coupling appears in the earliest decoder layer, while the strongest $\mathbf{C}_t\!\rightarrow\!\mathbf{H}_t$ coupling appears near the final decoder layers. Relative to the matched control, both couplings are substantially larger, indicating a structured interaction pattern rather than generic attention non-uniformity.

\paragraph{Support-region selectivity.}
We next test whether the \emph{scene-aware contact prompt} biases the human stream toward image regions that are physically relevant for support. For each frame, we project the ground-truth RICH contact vertices into the image plane to form a binary support-region mask, and define \emph{support selectivity} as the unitless ratio of average attention mass inside the support region to the average mass outside it. Fig.~\ref{fig:supp_internal_mechanism}\textbf{ (right)} plots this ratio across decoder layers for $\textrm{Human3R}^{\dagger}$, Parallel Readout, and \modelname. Averaged over the official dynamic-camera RICH test subset, the three models obtain $1.5$, $1.5$, and $1.7$, respectively, and \modelname\ exceeds both baselines in all camera-sequence entries. The gain is moderate but highly consistent, and indicates that the scene-aware contact prompt reallocates the human-stream attention toward support-relevant image regions.

\paragraph{Counterfactual residual ablation.}
Finally, we directly ablate the latent residual branch by setting the residual to zero at inference time and re-running only the final SMPL-X~\cite{SMPLX} regression head. This intervention changes the reconstructed body while leaving the contact prediction unchanged. Averaged over the moving-camera subset, the ablation induces a body-root translation change of $\|\Delta\boldsymbol{\Gamma}\|_2 = 0.3$\,m (here $\boldsymbol{\Gamma}$ denotes the world-frame SMPL-X root translation, i.e., $\mathbf{P}_t$ in the main paper notation), a mean per-joint angular shift of $2.8^\circ$, and a maximum per-joint shift of $10.2^\circ$, while the contact-logit difference remains $0.0$. This supports distinct roles for the two branches of the contact pathway: the contact head predicts dense contact, whereas the latent refinement branch uses the contact-conditioned representation to modify the final body configuration.

Taken together, these diagnostics support the same conclusion as the main experiments. In \modelname, contact is not used as a detached auxiliary output. Instead, it is routed through a dedicated contact representation and then fed back into the human stream through contact-guided latent refinement, where it acts as an internal corrective signal for the final body estimate.

\subsection{Comprehensive Comparison for Global Human Motion Estimation}
\label{ssec:hmr_full}

\cref{tab:global_hmr_full} provides the comprehensive comparison for global human motion estimation on RICH~\cite{huang2022capturing} and EMDB-2~\cite{kaufmann2023emdb}. In the main paper, we focus on the \emph{Unified} setting, where methods jointly recover humans, camera motion, and dense scene geometry from monocular video without per-sequence optimization. This is the setting most directly matched to \modelname. For completeness, we also include methods from two related settings.

We group methods by inference setting. \textbf{Offline} methods use per-sequence optimization or test-time optimization, and can therefore adapt predictions to the full test sequence. \textbf{World-HMR} methods estimate global human motion in world coordinates, but do not jointly reconstruct dense scene geometry in the same unified feed-forward model. \textbf{Unified} methods jointly predict humans, camera motion, and scene geometry in a feed-forward reconstruction pipeline. Since these categories differ in task scope and inference procedure, comparisons across categories should be interpreted as contextual rather than controlled.

Within the \textbf{Unified} category, \modelname\ achieves the best results on RICH across WA-MPJPE, W-MPJPE, RTE, and PA-MPJPE. On EMDB-2, it improves over all Human3R-derived baselines and achieves the best WA-MPJPE, RTE, and PA-MPJPE within the same category. Compared with $\textrm{Human3R}^{\dagger}$, the gains are not explained by the training protocol alone; compared with Parallel Readout, they show that contact is most useful when it is fed back into the reconstruction pathway rather than predicted as an auxiliary output.

The broader comparison also shows the remaining gap between unified feed-forward reconstruction and methods that use offline or post-hoc refinement. Some Offline and World-HMR methods obtain competitive global motion accuracy, but they do not address the same unified online human-scene reconstruction problem as \modelname. Our focus is therefore on improving global motion and physical grounding within a single feed-forward human-scene reconstruction pipeline.

\begin{table*}[t]
\centering
\caption{\textbf{Comprehensive Comparison for Global human motion estimation on RICH~\cite{huang2022capturing} and EMDB-2~\cite{kaufmann2023emdb}.}
\textbf{Scene} denotes 3D scene reconstruction, and \textbf{Contact} denotes dense human-scene contact prediction or use. $\textrm{Human3R}^{\ast}$: released checkpoint in our pipeline; $\textrm{Human3R}^{\dagger}$: same architecture with our training protocol. \textbf{Bold} and \underline{underline} indicate the best and second-best results within the \emph{Unified} category, where \modelname\ operates. WA-MPJPE, W-MPJPE, and PA-MPJPE are reported in mm; RTE is reported in \%.}
\label{tab:global_hmr_full}
\footnotesize
\tabstyle
\setlength{\tabcolsep}{2pt}
\begin{adjustbox}{width=\textwidth}
\begin{tabular}{@{}llcccccccccc@{}}
\toprule
\multirow{2}{*}{\textbf{Category}} &
\multirow{2}{*}{\textbf{Method}} &
\multirow{2}{*}{\textbf{Scene}} &
\multirow{2}{*}{\textbf{Contact}} &
\multicolumn{4}{c}{\textbf{RICH}~\cite{huang2022capturing}} &
\multicolumn{4}{c}{\textbf{EMDB-2}~\cite{kaufmann2023emdb}} \\
\cmidrule(lr){5-8} \cmidrule(lr){9-12}
& & & &
\textbf{WA-MPJPE} $\downarrow$ &
\textbf{W-MPJPE} $\downarrow$ &
\textbf{RTE (\%)} $\downarrow$ &
\textbf{PA-MPJPE} $\downarrow$ &
\textbf{WA-MPJPE} $\downarrow$ &
\textbf{W-MPJPE} $\downarrow$ &
\textbf{RTE (\%)} $\downarrow$ &
\textbf{PA-MPJPE} $\downarrow$ \\
\midrule

\multirow{3}{*}{\textbf{Offline}}
& SLAHMR~\cite{Ye_2023_CVPR_slamhr} & \xmark & \xmark & 132.2 & 237.1 & 6.4 & -- & 326.9 & 776.1 & 10.2 & 61.5 \\
& TRAM~\cite{TRAM} & \xmark & \xmark & 127.8 & 238.0 & 6.0 & -- & 76.4 & 222.4 & 1.4 & 38.1 \\
& JOSH~\cite{JOSH3R} & \cmark & \cmark & 102.6 & 184.3 & 3.4 & -- & 68.9 & 174.7 & 1.3 & -- \\
\midrule

\multirow{4}{*}{\textbf{World-HMR}}
& TRACE~\cite{TRACE} & \xmark & \xmark & 238.1 & 925.4 & 101.4 & -- & 529.0 & 1702.3 & 17.7 & 58.0 \\
& WHAM~\cite{WHAM} & \xmark & \xmark & 108.4 & 190.1 & 4.5 & -- & 135.6 & 354.8 & 6.0 & 38.2 \\
& GVHMR~\cite{GVHMR} & \xmark & \xmark & 78.8 & 126.3 & 2.4 & -- & 111.0 & 276.5 & 2.0 & -- \\
& OnlineHMR~\cite{zhao2026onlinehmr} & \cmark & \xmark & -- & -- & -- & -- & 93.5 & 310.4 & 2.2 & 40.1 \\
\midrule

\multirow{6}{*}{\textbf{Unified}}
& JOSH3R~\cite{JOSH3R} & \cmark & \xmark & 190.4 & 334.9 & 6.3 & -- & 220.0 & 661.7 & 13.1 & -- \\
& UniSH~\cite{li2026unish} & \cmark & \xmark & 118.1 & 183.2 & 4.8 & 50.0 & 118.5 & \textbf{270.1} & 5.8 & 44.4 \\
& $\textrm{Human3R}^{\ast}$~\cite{chen2025human3r} & \cmark & \xmark & 109.9 & 184.0 & 3.3 & 48.6 & \underline{118.2} & 286.1 & \underline{2.4} & \underline{38.7} \\
& $\textrm{Human3R}^{\dagger}$~\cite{chen2025human3r} & \cmark & \xmark & \underline{97.5} & \underline{153.2} & \underline{2.3} & \underline{41.4} & -- & -- & -- & -- \\
\addlinespace[2pt]
& \cellcolor{gray!10}\textbf{Parallel Readout (Ours)} & \cellcolor{gray!10}\cmark & \cellcolor{gray!10}\cmark & \cellcolor{gray!10}97.9 & \cellcolor{gray!10}153.5 & \cellcolor{gray!10}\underline{2.3} & \cellcolor{gray!10}41.5 & \cellcolor{gray!10}160.7 & \cellcolor{gray!10}388.3 & \cellcolor{gray!10}3.6 & \cellcolor{gray!10}50.9 \\
& \cellcolor{blue!5}\textbf{\modelname\ (Ours)} & \cellcolor{blue!5}\cmark & \cellcolor{blue!5}\cmark & \cellcolor{blue!5}\textbf{81.5} & \cellcolor{blue!5}\textbf{129.8} & \cellcolor{blue!5}\textbf{1.7} & \cellcolor{blue!5}\textbf{37.6} & \cellcolor{blue!5}\textbf{113.7} & \cellcolor{blue!5}\underline{285.6} & \cellcolor{blue!5}\textbf{2.3} & \cellcolor{blue!5}\textbf{38.2} \\
\bottomrule

\end{tabular}
\end{adjustbox}
\end{table*}

\subsection{Runtime Efficiency}
\label{ssec:efficiency}

In all our experiments, $\textrm{Human3R}^{\ast}$ denotes our evaluation of the released Human3R checkpoint (ViT-L/896 variant) under the same preprocessing, inference, and evaluation pipeline used for \modelname. For end-to-end reconstruction, $\textrm{Human3R}^{\ast}$~\cite{chen2025human3r} runs at 2.41 FPS on a single NVIDIA A5000 GPU, while \modelname\ runs at 2.40 FPS under the same preprocessing and benchmarking protocol. This shows that the contact branch adds negligible overhead to the full reconstruction pipeline.

For contact prediction, \modelname\ runs at 0.5 s/frame, compared with 1.1 s/frame for DECO~\cite{tripathi2023deco} and 3.5 s/frame for InteractVLM~\cite{dwivedi2025interactvlm}. Thus, \modelname\ is about $2.0\times$ faster than DECO and about $6.1\times$ faster than InteractVLM on this task. Unlike these methods, contact in \modelname\ is integrated into the reconstruction pipeline rather than applied as a separate post-processing stage.

\subsection{Backbone Scaling: Lightweight ViT-S/672 Variant}
\label{ssec:vits672}

The main paper instantiates \modelname\ on top of the largest Human3R~\cite{chen2025human3r} checkpoint (ViT-L/896 with Multi-HMR~\cite{Multi-HMR} ViT-L), which delivers the strongest accuracy at the cost of inference speed. Because Human3R itself is released as a family of backbones (ViT-S/672, ViT-B/672, ViT-L/672, ViT-L/896) that trade accuracy for speed, a natural question is whether the contact-aware design of \modelname\ continues to help when the underlying backbone is much smaller. We address this by also instantiating \modelname\ on the lightweight ViT-S/672 checkpoint, in which both the input resolution ($672\times 672$) and the Multi-HMR encoder size (ViT-S) are reduced. The two contributions of \modelname---the \emph{scene-aware contact prompt} and the \emph{contact-guided latent refinement}---are unchanged. The training protocol (RICH~\cite{huang2022capturing} + BEDLAM~\cite{black2023bedlam} fine-tuning, see \cref{ssec:impl}) is also identical to the main paper.

Under the lightweight configuration, \modelname\ runs at $5.64$\,FPS
on a single NVIDIA A5000 GPU (mean over the RICH moving-camera test
subset), compared with $5.85$\,FPS
for the corresponding Human3R~\cite{chen2025human3r} ViT-S/672 backbone
evaluated under the same protocol. The $\sim\!0.21$\,FPS overhead is
attributable to the additional contact branch and reinforces the
negligible-overhead behavior reported at the ViT-L/896 scale in
\cref{ssec:efficiency}. Compared with the ViT-L/896 \modelname\ from
the main paper ($2.40$\,FPS), the lightweight variant is therefore
approximately $2.4\times$ faster, making it the appropriate choice
when reduced compute budget is the primary requirement.

\paragraph{Merged comparison across backbones.}
\cref{tab:global_hmr_backbones} compares the two backbone variants of
\modelname\ side by side against backbone-matched Human3R baselines:
$\textrm{Human3R}^{\ast}$, the released checkpoint at each scale,
evaluated in our pipeline; and $\textrm{Human3R}^{\dagger}$, the same
architecture and backbone fine-tuned with our training protocol so
that supervision is matched. Because the two backbones differ
substantially in capacity, we apply the bold/underline convention
\emph{within each backbone group} rather than across groups.

\paragraph{Global human motion estimation.}
On global motion (\cref{tab:global_hmr_backbones}), the contact
pathway delivers consistent improvements at \emph{both} backbone
scales. At ViT-S/672, \modelname\ reduces WA-MPJPE on RICH from
$113.9$\,mm ($\textrm{Human3R}^{\dagger}$) to $97.2$\,mm ($-14.6\%$),
and W-MPJPE from $178.3$\,mm to $151.1$\,mm ($-15.3\%$). At
ViT-L/896, the corresponding reductions are $97.5\!\to\!81.5$\,mm
($-16.4\%$) on WA-MPJPE and $153.2\!\to\!129.8$\,mm ($-15.3\%$) on
W-MPJPE. The relative improvement from contact-aware refinement is
therefore stable across the $\sim\!2.4\times$ speed difference between
the two variants.

\paragraph{When to use which variant.}
Together, \cref{tab:global_hmr_backbones} supports a clean
recommendation. The lightweight ViT-S/672 variant is the appropriate
choice when compute budget is the primary constraint: it preserves the
world-coordinate gains of contact-guided refinement at $5.64$\,FPS,
roughly $2.4\times$ the throughput of the ViT-L/896 variant. For
applications where fine-grained body-mesh detail matters most---such
as expressive avatar reconstruction or close-range human-scene
interaction analysis---the ViT-L/896 variant from the main paper
remains the better choice. Importantly, the contact-aware design of
\modelname\ transfers cleanly across backbone scales without any
architecture change, indicating that the proposed contact pathway is
complementary to, rather than competing with, the backbone-scaling
axis explored by Human3R~\cite{chen2025human3r}.

\begin{table*}[t]
\centering
\caption{\textbf{Global human motion estimation across backbone scales on RICH~\cite{huang2022capturing} and EMDB-2~\cite{kaufmann2023emdb}.}
We compare the lightweight ViT-S/672 variant and the ViT-L/896 variant used in the main paper. \textbf{Scene} denotes 3D scene reconstruction; \textbf{Contact} denotes dense human-scene contact prediction or use. $\textrm{Human3R}^{\ast}$: released checkpoint at the indicated backbone, evaluated in our pipeline; $\textrm{Human3R}^{\dagger}$: same architecture and backbone with our training protocol. WA-MPJPE, W-MPJPE, and PA-MPJPE are in mm; RTE in \%. \textbf{Bold}/\underline{underline} denote best/second-best \emph{within each backbone group}.}
\label{tab:global_hmr_backbones}
\footnotesize
\setlength{\tabcolsep}{3pt}
\begin{adjustbox}{width=\textwidth}
\begin{tabular}{@{}llcccccccccc@{}}
\toprule
\multirow{2}{*}{\textbf{Method}} &
\multirow{2}{*}{\textbf{Backbone}} &
\multirow{2}{*}{\textbf{Scene}} &
\multirow{2}{*}{\textbf{Contact}} &
\multicolumn{4}{c}{\textbf{RICH}~\cite{huang2022capturing}} &
\multicolumn{4}{c}{\textbf{EMDB-2}~\cite{kaufmann2023emdb}} \\
\cmidrule(lr){5-8} \cmidrule(lr){9-12}
& & & &
\textbf{WA-MPJPE} $\downarrow$ &
\textbf{W-MPJPE} $\downarrow$ &
\textbf{RTE (\%)} $\downarrow$ &
\textbf{PA-MPJPE} $\downarrow$ &
\textbf{WA-MPJPE} $\downarrow$ &
\textbf{W-MPJPE} $\downarrow$ &
\textbf{RTE (\%)} $\downarrow$ &
\textbf{PA-MPJPE} $\downarrow$ \\
\midrule
$\textrm{Human3R}^{\ast}$~\cite{chen2025human3r} & ViT-S/672 & \cmark & \xmark & 131.5 & 207.3 & 3.3 & 71.8 & \underline{137.8} & \textbf{333.1} & \textbf{2.4} & \textbf{52.9} \\
$\textrm{Human3R}^{\dagger}$~\cite{chen2025human3r} & ViT-S/672 & \cmark & \xmark & \underline{113.9} & \underline{178.3} & \underline{2.6} & \underline{58.3} & -- & -- & -- & -- \\
\rowcolor{blue!5}\textbf{\modelname\ (Ours)} & ViT-S/672 & \cmark & \cmark & \textbf{97.2} & \textbf{151.1} & \textbf{2.0} &  \textbf{51.7} & \textbf{130.7} & \underline{331.6} & \underline{2.5} & \textbf{52.9} \\
\midrule
$\textrm{Human3R}^{\ast}$~\cite{chen2025human3r} & ViT-L/896 & \cmark & \xmark & 109.9 & 184.0 & 3.3 & 48.6 & \underline{118.2} & \underline{286.1} & \underline{2.4} & \underline{38.7} \\
$\textrm{Human3R}^{\dagger}$~\cite{chen2025human3r} & ViT-L/896 & \cmark & \xmark & \underline{97.5} & \underline{153.2} & \underline{2.3} & \underline{41.4} & -- & -- & -- & -- \\
\rowcolor{blue!5}\textbf{\modelname\ (Ours)} & ViT-L/896 & \cmark & \cmark & \textbf{81.5} & \textbf{129.8} & \textbf{1.7} & \textbf{37.6} & \textbf{113.7} & \textbf{285.6} & \textbf{2.3} & \textbf{38.2} \\
\bottomrule
\end{tabular}
\end{adjustbox}
\end{table*}

\section{Implementation Details}
\label{ssec:impl}

\paragraph{Backbone and initialization.}
We build \modelname\ on top of the released Human3R~\cite{chen2025human3r} checkpoint, which adapts CUT3R~\cite{cut3r} through visual prompt tuning. As in Human3R, we keep the pretrained CUT3R backbone frozen during fine-tuning, and we also keep the Multi-HMR~\cite{Multi-HMR} encoder frozen. The frozen scene backbone uses a ViT-Large image encoder with token dimension $1024$, followed by two interleaved 4D decoders inherited from CUT3R with embedding dimension $c{=}768$, depth $12$, and $12$ attention heads. The persistent state $\mathbf{S}_t$ also has dimension $768$.

\paragraph{Trainable modules.}
We optimize only the human- and contact-related modules introduced by \modelname. These include the scene-context cross-attention layers, the prompt-construction MLPs used to form the \emph{scene-aware contact prompt}, the contact head, the residual head $\mathrm{Head}_{\mathrm{residual}}$ used for \emph{contact-guided latent refinement}, the SMPL-X prediction head, and the segmentation head inherited from Human3R. Unless noted otherwise, all MLPs use two linear layers with GELU activations and hidden dimension $768$. The SMPL-X parameter decoders inside $\mathrm{Head}_{\mathrm{human}}$ follow Human3R and use separate two-layer MLPs with hidden dimension $4\times 1792$ to regress body pose, shape, expression, and root translation from $\bar{\mathbf{H}}_t$.

\paragraph{Local geometry token.}
The explicit metric geometry cue in Sec.~\ref{ssec:contact_prompt} is constructed from the previous-frame world-frame pointmap $\mathbf{X}_{t-1}$ using the same notation as in the main paper. For each detected human anchor $\mathbf{u}_t^n$, we apply RoIAlign~\cite{he2017mask} to a square window centered at $\mathbf{u}_t^n$ with half-size $24$ pixels and output size $7\times7$. The pooled $3$-channel coordinates are then averaged to obtain the local descriptor $\phi_{\mathrm{geo}}(\mathbf{X}_{t-1}, \mathbf{u}_t^n)\in\mathbb{R}^3$, which is mapped into the decoder space to form the geometry token used in $\mathbf{G}_t$.

\paragraph{Training data.}
We fine-tune on a balanced mixture of RICH~\cite{huang2022capturing} and BEDLAM~\cite{black2023bedlam}. Following CUT3R and Human3R, we exclude BEDLAM sequences whose environment is represented as a panoramic HDRI image, yielding $2700$ training sequences. RICH provides multi-camera video with scanned scene geometry, SMPL-X meshes, and dense body-scene contact annotations; we use the official split of RICH sequences leaving the moving camera subset for evaluation.

\paragraph{Optimization.}
Training uses AdamW with $\beta_1{=}0.9$, $\beta_2{=}0.95$, and weight decay $0.05$. We use a linear warmup over the first $5$ epochs from $10^{-6}$ to $10^{-5}$, followed by cosine decay back to $10^{-6}$, and train for $100$ epochs. The per-GPU batch size is $4$ with gradient accumulation of $2$, giving an effective batch size of $64$ on $8$ NVIDIA A5000 GPUs. We use mixed precision and gradient checkpointing.

\paragraph{Evaluation Metrics}
For evaluating global human motion, we split sequences into segments of 100 frames
and align the predicted motion with the ground truth using either the first two frames or the entire segment and compute the mean-per-joint error in millimeters to obtain W-MPJPE or WA-MPJPE
for evaluation. We also evaluate the relative Root Translation Error (RTE\%) of the whole trajectory.

\paragraph{Training objective.}
The total training objective is
\[
\mathcal{L}_{\mathrm{total}}
=
\mathcal{L}_{\mathrm{4D}}
+
\mathcal{L}_{\mathrm{SMPLX}}
+
\lambda_c \mathcal{L}_{\mathrm{contact}}.
\]
For the inherited reconstruction pathway, we use the CUT3R confidence-aware $L_{2,1}$ pointmap loss with $\alpha{=}0.2$ together with the per-frame RGB MSE loss, and supervise the human output with the Human3R $L_1$ SMPL-X loss. For contact, we use focal BCE with $\gamma{=}2.0$ and $\alpha{=}0.25$, a positive-class weight of $10$, and a part-level loss weighted by $\lambda_p{=}1.5$, with overall contact weight $\lambda_c{=}1.0$. Since BEDLAM does not provide body-scene contact labels, contact supervision is applied only on RICH samples, while BEDLAM samples optimize only the inherited reconstruction and HMR objectives.

\paragraph{Contact supervision space.}
Following BSTRO~\cite{huang2022capturing}, we supervise contact in SMPL~\cite{SMPL} space rather than SMPL-X space. The human regressor remains in SMPL-X space for compatibility with Human3R, while the contact head predicts dense contact on the $6890$ SMPL vertices, where most body-scene support contact occurs.

\section{Physical Plausibility: Qualitative Examples}
\label{ssec:phys_eval_fig}

\begin{figure*}[h]
    \centering
    \includegraphics[width=\linewidth]{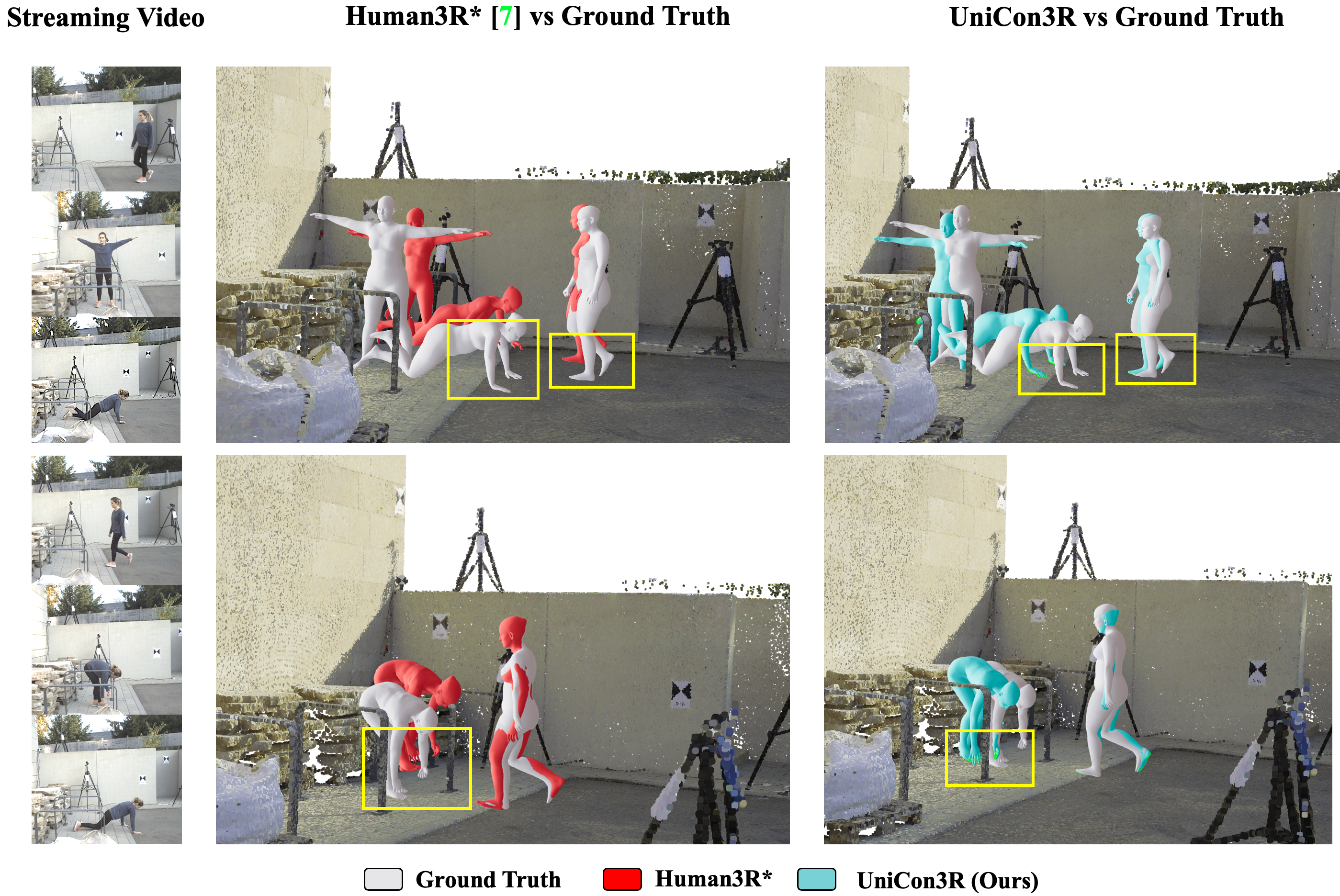}  %
    \caption{\textbf{Qualitative comparison of physical plausibility on RICH~\cite{huang2022capturing}.}
    We compare Human3R~\cite{chen2025human3r} and \modelname\ against ground truth on a streaming sequence. Yellow boxes highlight floating or implausible 4D reconstructions of Human3R compared to the more grounded predictions of \modelname.}
    \label{fig:supp_phys_qual}
\end{figure*}

We complement the quantitative grounding metrics of \cref{ssec:phys_eval}
(Penetrate, Float, Pen.~Max, Collision Ratio in Table~\ref{tab:contact_grounding})
with qualitative examples on the moving-camera subset of RICH~\cite{huang2022capturing}.
\cref{fig:supp_phys_qual} compares Human3R~\cite{chen2025human3r} (red)
and \modelname\ (cyan) against the ground-truth body (gray) on a streaming
sequence. Yellow boxes highlight regions where Human3R produces visibly
floating or penetrating reconstructions: the body either hovers above the
support surface or intersects the scene mesh. \modelname\ places the body
closer to the support surface in these same frames, producing more
plausible body--scene interaction. The visual pattern matches the
quantitative reductions reported in \cref{ssec:phys_eval} (Penetrate
$6.40 \to 1.59$\,cm, Float $33.56 \to 5.50$\,cm, Pen.~Max $216.97 \to 17.54$\,cm)
and shows that the contact-guided refinement of \modelname\ resolves the
specific failure modes that drive Human3R's grounding errors.

\newpage

\end{document}